\documentclass[10pt,journal,compsoc]{IEEEtran}
\usepackage{amssymb}
\usepackage{amsmath}
\usepackage{float}
\usepackage[figuresright]{rotating}
\usepackage{array}
\usepackage{graphicx}
\usepackage{multirow}
\usepackage[table,xcdraw]{xcolor}
\ifCLASSOPTIONcompsoc
  \usepackage[nocompress]{cite}
\else
  \usepackage{cite}
\fi
\ifCLASSINFOpdf
\else
\fi
\begin{document}
%
\title{A Review on Facial Micro-Expressions Analysis: Datasets, Features and Metrics}
%
%
%
%

\author{Walied~Merghani,
	   ~Adrian~K.~Davison, ~\IEEEmembership{Member,~IEEE},
       ~Moi~Hoon~Yap, ~\IEEEmembership{Member,~IEEE} \\ (\textit{This work has been submitted to the IEEE for possible publication. Copyright may be transferred without notice, after which this version may no longer be accessible}).
\IEEEcompsocitemizethanks{\IEEEcompsocthanksitem W. Merghani is with Sudan University of Science and Technology
\IEEEcompsocthanksitem A. K. Davison was with Manchester Metropolitan University during this study and is now with the University of Manchester.
\IEEEcompsocthanksitem M. H. Yap is with the School of Computing, Mathematics and Digital Technology, Manchester Metropolitan University, Manchester, United Kingdom. E-mail: m.yap@mmu.ac.uk}
}

%
%

\markboth{Preprint submitted to IEEE Journal.}%
{Shell \MakeLowercase{\textit{et al.}}: A Review on Facial Micro-expressions}
%



\IEEEtitleabstractindextext{%
\begin{abstract}
Facial micro-expressions are very brief, spontaneous facial expressions that appear on the face of humans when they either deliberately or unconsciously conceal an emotion. Micro-expression has shorter duration than macro-expression, which makes it more challenging for human and machine. Over the past ten years, automatic micro-expressions recognition has attracted increasing attention from researchers in psychology, computer science, security, neuroscience and other related disciplines. The aim of this paper is to provide the insights of automatic micro-expressions analysis and recommendations for future research. There has been a lot of datasets released over the last decade that facilitated the rapid growth in this field. However, comparison across different datasets is difficult due to the inconsistency in experiment protocol, features used and evaluation methods. To address these issues, we review the datasets, features and the performance metrics deployed in the literature. Relevant challenges such as the spatial temporal settings during data collection, emotional classes versus objective classes in data labelling, face regions in data analysis, standardisation of metrics and the requirements for real-world implementation are discussed. We conclude by proposing some promising future directions to advancing micro-expressions research.

\end{abstract}

}

\maketitle

\IEEEdisplaynontitleabstractindextext

%
\IEEEpeerreviewmaketitle

\IEEEraisesectionheading{\section{Introduction}\label{sec:introduction}}

%
%
%
%

\IEEEPARstart{F}{acial} expression research has a long history and accelerated through the 1970s. The modern theory on basic emotions by Ekman et al~\cite{ekman1992argument,Ek04,ekman2005what} has generated more research than any other in the psychology of emotion~\cite{russell1997psychology}. They outline 7 universal facial expressions: happy, sad, anger, fear, surprise, disgust and contempt, as the universality of emotion. When an emotional episode is triggered, there is an impulse which may induce one or more of these expressions of emotion.

Facial micro-expression (henceforth, micro-expression) analysis has become an active research area in recent years. Micro-expressions occur when a person attempts to conceal their true emotion~\cite{Ek04,Ek09}. When they consciously realise that a facial expression is occurring, the person may try to suppress the facial expression because showing the emotion may not be appropriate or could be due to a cultural display rule~\cite{matsumoto2008culture}. Once the suppression has occurred, the person may mask over the original facial expression and cause a micro-expression. In a high-stakes environment, these expressions tend to become more likely as there is more risk to showing the emotion. 

Micro-expressions contain a significant and effective amount of information about the true emotions which may be useful in practical applications such as security and interrogations~\cite{osullivan2009police,frank2009behavior,frank2009see}. It is not easy to extract this information due to the brief movements in micro-expressions, where there is a need for the features to be more descriptive. The difficulty also comes from one of the main characteristics of micro-expressions which is the short duration, with the general standard being a duration of no more than 500 ms~\cite{yan2013how}. Other definitions of speed that have been studied show micro-expressions to last less than 250 ms~\cite{ekman2001telling}, less than 330 ms~\cite{ekman2005what} and less than half a second~\cite{frank2009behavior}. Following Ekman and Friesen as first to define a micro-expression~\cite{ekman1969nonverbal}, a usual duration considered is less than 200 ms. Duration is the main feature that distinguishing micro-expressions from macro-facial expressions ~\cite{shreve2012effects}, which make it more challenging than micro-expressions in the following aspects:
\begin{itemize}
   \item {Difficulties for human to spot micro-expressions: Humans find it difficult to spot micro-expressions consistently~\cite{frank2009behavior}. This is due to macro-expressions tend to be large and distinct, whereas micro-expressions are very quick and subtle muscle movements.}
   \item {Datasets Creation: It is difficult to induce micro-expressions if compared to macro-expressions. Current available micro-expression datasets were induced in a laboratory controlled environment. Macro-expressions can be recorded by normal camera. However, the speed and subtlety of micro-expressions require high-speed camera, where this digital capture device produces more noisy data than the normal camera. }
   \item {The history of algorithm development: Automated micro-expression recognition is relatively new (found work in 2009 \cite{polikovsky2009facial}), when compared with facial expression recognition (found in 1990s \cite{essa1997coding, kimura1997facial}).}
\end{itemize}

Although both micro and macro-expressions loosely related due to the facial expression aspect, these two topics should be looked upon as different research problems. Our focus is to provide comprehensive review and new insights for micro-expressions. For review in macro-expressions, please refer to \cite{pantic2000automatic, fasel2003automatic}.

This paper introduces and surveys recent research advances of micro-expressions. We present a comprehensive review and comparison on the datasets, the state-of-the-art features for micro-expression recognition and the performance metrics.  We demonstrate the potential and challenges of micro-expression analysis. This rest of the paper is organised as follows: Section 2 provides a review on publicly available datasets. Section 3 presents the feature representation. Detailed performance metrics used in this field are shown in Section 4 and Section 5 outlines challenges and Section 6 concludes this paper by providing future recommendations.

\section{Facial Micro-expression Datasets}
\label{sec:lit}
This section will compare and contrast the relevant publicly available datasets for facial micro-expressions analysis.

\subsection{Non-spontaneous datasets}
The earlier research dependent on non-spontaneous datasets. Here we present a review on the three earliest non-spontaneous datasets.
\subsubsection{Polikovsky Dataset}
One of the first micro-expression datasets was created by Polikovsky et al.~\cite{polikovsky2009facial}. The participants were 10 university students in a laboratory setting and their faces were recorded at 200fps with a resolution of 640$\times$480. The demographic was reasonably spread but limited in number with 5 Asians, 4 Caucasians and 1 Indian student participants.

The laboratory setting was set up to maximise the focus on the face, and followed the recommendations of mugshot best practices by McCabe~\cite{mccabe2009best}. To reduce shadowing, lights were placed above, to the left and right of the participant. The background consisted of a uniform colour of approximately 18\% grey. The camera was also rotated 90 degrees to increase the pixels available for face acquisition.

The micro-expressions in this dataset were posed by participants whom were asked to perform the 7 basic emotions with low muscle intensity and moving back to neutral as fast as possible. Posed facial expressions have been found to have significant differences to spontaneous expressions~\cite{afzal2009natural}, therefore the micro-expressions in this dataset are not representative of natural human behaviour and highlights the requirement for expressions induced naturally. Further, this dataset is not publicly available for further study.

\subsubsection{USF-HD}
Similar to the previous dataset, USF-HD~\cite{shreve2011macro} includes 100 posed micro-expressions recorded at 29.7 fps. The participants were shown various micro-expressions and told to replicate them in any order of preferencethey ed. As with the Polikovly described dataset, posed  not re-create a real-world scenario and replicating other people's micro-expressions does not represent how these movements would be presented by the participants themselves.

Recording at almost 30 fps can risk losing important information about the movements. In addition, this dataset defined micro-expressions as no higher than 660 ms, which is longer than the previously accepted definitions. Moreover, the categories for micro-expressions are smile, surprise, anger and sad, which is reduced from the 7 universal expressions by missing out disgust, fear and contempt. This dataset has also not been made available for public research use.

\subsubsection{YorkDDT}
As part of a psychological study named the York Deception Detection Test (YorkDDT), Warren et al.~ \cite{warren2009detecting} recorded 20 video clips, at 320$\times$240 resolution and 25 fps, where participants truthfully or deceptively described two film clips that were either classed as emotional, or non-emotional. The emotional clip, intended to be stressful, was of an unpleasant surgical operation. The non-emotional clip was meant to be neutral, showing a pleasant landscape scene.

The participants viewing the emotional clip were asked to describe the non-emotional video, and vice versa for the participants watching the non-emotional clip. Warren et al.~\cite{warren2009detecting} reported that some micro-expressions occurred during both scenarios, however these movements were not reported to be available for public use.

During their study into micro-expression recognition, Pfister et al.~\cite{pfister2011recognising} managed to obtain the original scenario videos where 9 participants (3 male and 6 female) displayed micro-expressions. They extracted 18 micro-expressions for analysis, 7 from the emotional scenario and 11 from the non-emotional version.

Other than the very low amount of micro-expressions in this dataset, it is created through a second source that do not go into a large amount of detail about AU, or participant demographic. With the data unable to be publicly accessed, it is not possible to study these micro-expressions. It is also an issue with the frame rate being so low, the largest amount of frames for analysis would be around 12-13 frames. The lowest reported micro-expression length was 7 frames.

\begin{figure*}[]
	\centering
	\includegraphics[scale=0.5]{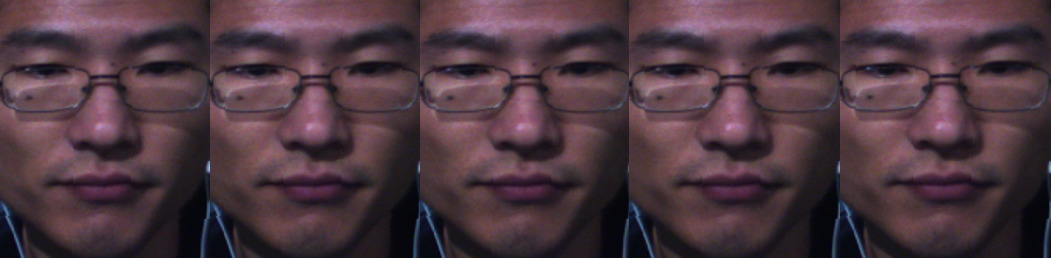}
	\centering
	\caption{Sample of HS SMIC dataset with negative expression.}
	\label{SMICFIG}
\end{figure*}

\begin{figure*}[]
	\centering
	\includegraphics[scale=0.255]{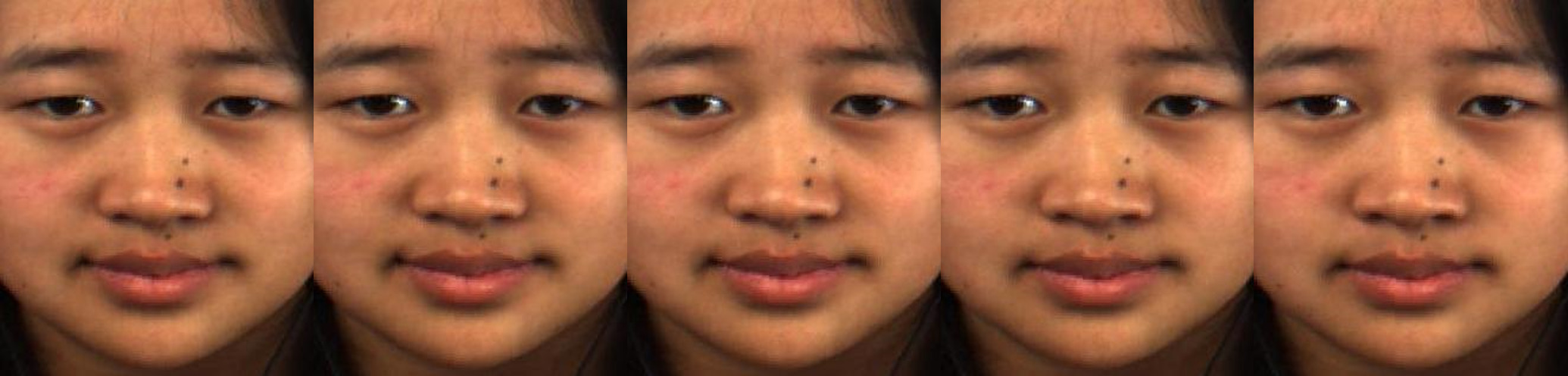}
	\caption{Sample of CASME II dataset with happiness expression, the participant has been FACS coded with AU1+AU12 (Inner brow raiser+lip corner puller).}
	\label{CASMEFIG}
\end{figure*}

\begin{figure*}[]
	\centering
	\includegraphics[scale=0.2]{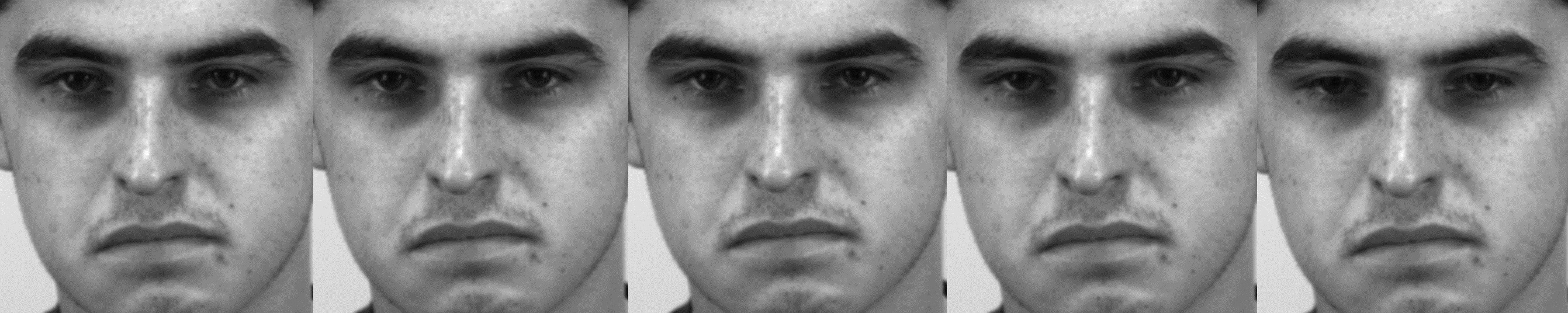}
	\caption{Sample of SAMM dataset with anger expression, the participant has been
		FACS coded with AU4+AU7 (Brow lowerer+lid tightener).}
	\label{SAMMFIG}
\end{figure*}

\subsection{Spontaneous datasets}
Developing micro-expression spontaneous datasets is one of the biggest challenges faced in this research area. It is difficult to elicit micro-expressions because they are difficult to fake, so we need to get the true emotion while the person try to hide it. Some spontaneous datasets to date include: SMIC \cite{li2013spontaneous}, CASME \cite{yan2013casme}, CASME II \cite{yan2014casme}, SAMM \cite{davison2018samm} and CAS(ME)$^2$ \cite{qu2017cas}. SAMM was designed for micro-movements with less emphasis on the emotional side for increased objectiveness. Available datasets will be described in this section.

\subsubsection{Chinese Academy of Sciences Micro-Expressions (CASME)}
Yan et al. \cite{yan2013casme} created a spontaneous micro-expression dataset called CASME. The dataset contains 195 samples of micro-expressions with a frame rate of 60 fps. These 195 samples were selected from more than 1500 facial movements, where 35 participants (13 females, 22 males) took part. The clips were divided into two classes depending on the environmental setting and cameras used.
\paragraph{Class A}
Samples in this class recorded by BenQ M31 camera at 60 fps, and the resolution is set to 1280$\times$720 pixels. Natural light was used for recording.
\paragraph{Class B}
A GRAS-03K2C camera recording at 60 fps was used to record samples in this class with resolution set to 640$\times$480 pixels. For class B two LED lights have been used.

Table \ref{tab:casme} shows each emotion class and the frequencies at which they occur in the CASME(A and B) dataset.
\begin{table}[H]
	\centering
	\caption{The frequency occurance of each emotion category in the CASME dataset}
	\label{tab:casme}
	\renewcommand{\arraystretch}{1.2}
	\begin{tabular}{ | l | l | }
		\hline
		Emotion & Frequency \\ \hline
		Amusement & 5 \\ \hline
		Sadness & 6 \\ \hline
		Disgust & 88 \\ \hline
		Surprise & 20 \\ \hline
		Contempt & 3 \\ \hline
		Fear & 2 \\ \hline
		Repression & 40 \\ \hline
		Tense & 28 \\ \hline
	\end{tabular}
\end{table} 

\subsubsection{Spontaneous Micro-expression Corpus (SMIC)}
Li et al. \cite{li2013spontaneous} built the SMIC dataset, which was recorded in an indoor environment with four lights from the four upper corners of the room. To induce strong emotions 16 movie clips were selected and shown to participants on a computer monitor. Facial expressions have been gathered using a camera fixed on the top of monitor while participants watched movie clips.

The dataset is spontaneous, 20 participants (6 females and 14 males) participated in the experiment. A high speed (HS) camera set to 100 fps and resolution of 640$\times$480 was used to gather the expressions from the first ten participants. A sample from this HS dataset is shown in Fig. \ref{SMICFIG}. A normal visual camera (VIS) and near-infrared (NIR), both with 25 fps and resolution of 640$\times$480, were used for all 20 participants. The lower frame rates of the latter two cameras can help to check whether the current method can be effective at this speed.

The accepted duration of micro-expression for SMIC is 1/2 second. Since not every participant showed micro-expressions when recording SMIC the final dataset includes 164 micro-expression clips from 16 participants recorded in HS dataset. While VIS and NIR datasets include 71 clips from 8 participants. Emotions in SMIC were classified into 3 classes (positive, negative and surprise). Table \ref{smic} show the number of emotions in any class according to the type of dataset.

\begin{table}[]
	\centering
	\caption{Type of Emotions Frequency in SMIC \cite{li2013spontaneous}}
	\label{smic}
	\renewcommand{\arraystretch}{1.2}
	\begin{tabular}{|c|c|c|c|c|} \hline
		Dataset & positive & negative & surprise & total \\ \hline
		HS      & 51       & 70       & 43       & 164   \\ \hline
		VIS     & 28       & 23       & 20       & 71    \\ \hline
		NIR     & 28       & 23       & 20       & 71    \\	\hline
	\end{tabular}
\end{table}

\subsubsection{Chinese Academy of Sciences Micro-Expression II (CASME II)}
CASME II has been developed by Yan et al.~\cite{yan2014casme}, which succeeds the CASME dataset \cite{yan2013casme} with major improvements. All samples in CASME II are spontaneous and dynamic micro-expressions with high frame rate (200 fps). There is always a few frames kept before and after each micro-expressions, to make it suitable for detection experiments, however the amount of these frames can vary across clips. The resolution of samples is 640$\times$480 pixels for recording, which were saved as MJPEG with a resolution of around 280$\times$340 pixels for the cropped facial area. Fig. \ref{CASMEFIG} shows a sample from the CASME II with a happiness-class expression. The micro-expressions were elicited in a well-controlled laboratory environment. The dataset contains 247 micro-expressions (gathered from 35 participants) that were selected from nearly 3000 facial movements and have been labeled with action units (AUs) based on the Facial Action Coding System (FACS) \cite{ekman1978facs}. Lighting flickers were avoided in the recordings and highlights to the regions of the face have been reduced.

\begin{table}[H]
	\centering
	\caption{The frequency of each micro-expression class in the CASME II dataset.}
	\label{casme2}
	\renewcommand{\arraystretch}{1.2}
	\begin{tabular}{|l|l|}
		\hline
		Emotion                       & Frequency
		\\ \hline 
		Happiness                     & 33                             \\ \hline
		Disgust                       & 60                             \\ \hline
		Surprise                      & 25                             \\ \hline
		Repression                    & 27                             \\ \hline
		Others                        & 102                            \\ \hline
	\end{tabular}
\end{table}

\subsubsection{Spontaneous Actions and Micro-Movements (SAMM)}
The Spontaneous Actions and Micro-Movements (SAMM) \cite{davison2018samm} dataset is the first high-resolution dataset of 159 micro-movements induced spontaneously with the largest variability in demographics. The inducement procedure was based on the 7 basic emotions \cite{Ek04} and recorded at 200 fps. An example from the SAMM dataset can be seen in Fig. \ref{SAMMFIG}. As part of the experimental design, each video stimuli was tailored to each participant, rather than getting self-reports after the experiment. This allowed for particular videos to be chosen and shown to participants for optimal inducement potential. The experiment comprised of 7 stimuli used to induce emotion in the participants who were told to suppress their emotions so that micro-movements might occur. To increase the chance of this happening, a prize of £50 was offered to the participant that could hide their emotion the best, therefore introducing a high-stakes situation~\cite{Ek04,Ek09}. Each participant completed a questionnaire prior to the experiment so that the stimuli could be tailored to each individual to increase the chances of emotional arousal. There is a total of 159 FACS-coded micro-movements reported in this dataset.



\subsubsection{A Dataset of Spontaneous Macro-Expressions and Micro-Expressions (CAS(ME)$^2$)}
Qu et al. \cite{qu2017cas} presented a new facial database with macro- and micro-expressions, which included 250 and 53 samples respectively selected from more than 600 facial movements. This database has been collected from 22 participants (6 males and 16 females) with mean age of 22.59 years (standard deviation: 2.2). A Logitech Pro C920 camera was used to record samples at frame rate equal to 30 fps and resolution set to 640$\times$480 pixels. CAS(ME)$^2$ has been labelled using combinations of AUs, self-reports and the emotion category decided for the emotion-evoking videos. This database contains four emotion categories: positive, negative, surprise and other which is shown in Table \ref{qu2016} with their frequency occurrence.

\begin{table}[H]
	\centering
	\caption{Type of emotion and their frequencies in the CAS(ME)$^2$ dataset.}
	\label{qu2016}
	\renewcommand{\arraystretch}{1.2}
\begin{tabular}{ | l | l | l | }
	\hline
	Emotion & Macro-expression & Micro-expression \\ \hline
	Positive & 87 & 6 \\ \hline
	Negative & 95 & 19 \\ \hline
	Surprise & 13 & 9 \\ \hline
	Other & 55 & 19 \\ \hline
\end{tabular}
\end{table}        

\begin{table*}[h]
	\caption{A Summary of non-spontaneous and spontaneous datasets.}
	\label{data-comp}
	\centering
	\renewcommand{\arraystretch}{1.2}
	\begin{tabular}{|c|c|c|c|c|c|c|c|c|}
		\hline
		Dataset   & Participants & Resolution  & FPS  & Samples  & Emotion Classes & FACS Coded & Ethnicities \\ \hline
		Polikovsky~\cite{polikovsky2009facial}  & 11 & 640$\times$480   & 200    & 13  & 7    & Yes & 3  \\ \hline
		USF-HD~\cite{shreve2011macro}  & N/A  & 720$\times$1280  & 29.7  & 100   & 4 & No  & N/A  \\ \hline
		YorkDDT~\cite{warren2009detecting}  & 9  & 320$\times$240  & 25  & 18 & N/A  & No  & N/A \\ \hline
		CASME~\cite{yan2013casme} & 35  & 640$\times$480, 1280$\times$720 & 60  & 195   & 7 & Yes  & 1  \\ \hline
		SMIC~\cite{li2013spontaneous} & 20  & 640$\times$480  & 100 and 25 & 164  & 3 & No & 3 \\ \hline
		CASME II~\cite{yan2014casme}  & 35 & 640$\times$480 & 200 & 247  & 5 & Yes   & 1\\ \hline
		SAMM~\cite{davison2018samm}   & 32 & 2040$\times$1088 & 200 & 159  & 7 & Yes & 13          \\ \hline
		CAS(ME)$^2$~\cite{qu2017cas}  & 22 & 640$\times$480 & 30 & 250 macro, 53 micro & 4 & No & 1\\ \hline
	\end{tabular}
\end{table*}

\subsection{Dataset comparison}
Table~\ref{data-comp} shows summary of a comparison of the datasets. Due to the non-spontaneous datasets were not made available, it is not been possible to provide a critical review on those datasets. Overall, CASME II has a high number of micro-expression samples collected from high number of participants (35 participants), similar to CASME but with 195 samples. There is no distribution in ethnicities in CASME and CASME II, where all participants are Chinese. SMIC have participants from 3 different ethnicities, but this limitation was overcome in SAMM which has participants from 13 different ethnicities. SAMM also has advantage over the other in age distribution with mean age of 33.24 years (SD: $\pm$11.32). CASME II and SAMM have high frame rate (200 fps). SAMM is the first high-resolution dataset which set to 2040$\times$1088 pixel and a facial area of 400$\times$400. The CAS(ME)$^2$ has a limited number of micro-expression samples with just 53 collected. In terms of emotion stimuli for the participants, CASME and SAMM have 7 classes, CASME II has 5 classes and SMIC only with 3 classes. CASME, CASME II and SAMM have been coded using FACS. Although SAMM was stimulated by 7 emotional classes, the final label in their first release for the micro-movements only consists of FACS codes - but not emotion classes.

 CASME II and SAMM become the focus of the researchers as they equipped with all the criteria needed for micro-expressions recognition: emotion classes, high frame rate, a rich number of micro-expressions and varies in term of the  intensity for facial movements. 

\section{Features}
\begin{figure*}
	\centering
	\includegraphics[scale=0.65]{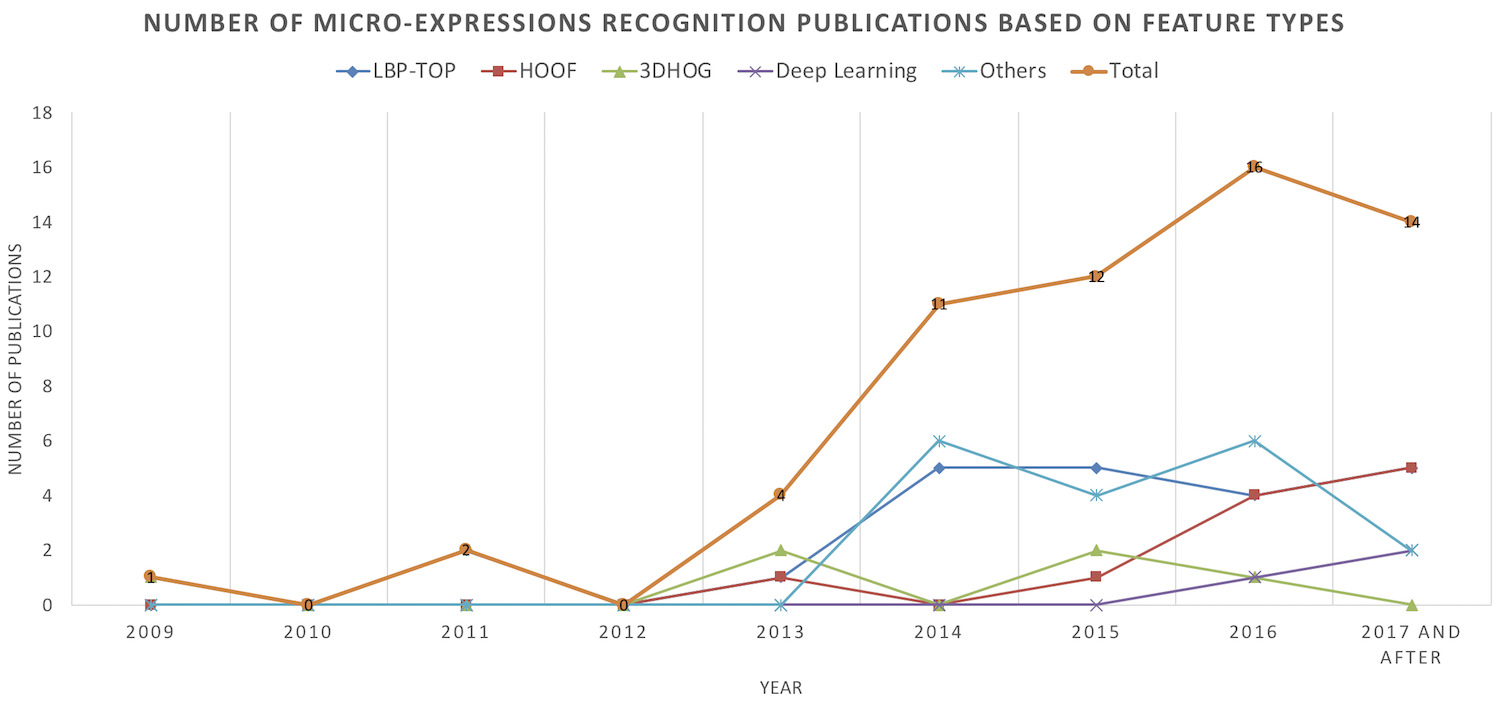}
	\caption{Illustration of the number of publications in micro-expressions recognition based on feature types over the past 10 years. The total publications show the growth of research in this area particularly in 2014 with more researchers focused on LBP-TOP. In recent years, more research focused on HOOF than 3DHOG. Deep learning technique is still in its infancy, but it is expected that the number of publications using deep learning will have a rapid growth in the future.}
	\label{fig:numPub}
\end{figure*}

The features used in micro-expression recognition will be discussed in this section. Figure \ref{fig:numPub} shows the total number of publications and its feature types based on our review. This is a strong evidence to support the growth of micro-expressions research. It is noted that 3DHOG was used in earlier work but not as popular as HOOF in recent years. LBP-TOP gained popularity in 2014 and maintained its number till today. On the other hand, deep learning is still in its infancy but we expect this number will grow rapidly in the future.

The full summary of the feature types, classifers and metrics used in the past decade is presented on Table \ref{table:Related-work} for Part I (2009-2015) and Table \ref{table:Related-workII} for Part II (2016-2018). The detailed algorithms review are categorised into: 3D Histograms of Oriented Gradients, Local Binary Pattern-Three Orthogonal Planes, Histogram of Oriented Optical Flow, Deep Learning Approaches and Other Feature Extraction Methods.

\subsection{3D Histograms of Oriented Gradients (3DHOG)}
Polikovsky et al. \cite{polikovsky2009facial} presented an approach for facial micro-expression recognition. They divided face into 12 regions selected through manual annotation of points on the face and then a rectangle was centred on these points. 3D-histograms of oriented gradients (3DHOG) was used to recognise motion in each region. This approach was evaluated on a posed dataset of micro-expressions captured using a high speed camera (200 fps). 13 different micro-expressions were recognised in this experiment. Their main contribution was to measure the duration of three phases of micro-expressions; constrict of the muscles (Constrict), muscle construction (In-Action) and release of the muscles (Release).

Polikovsky and Kameda~\cite{polikovsky2013facial} used 3DHOG again this time with k-mean classifier and voting procedure. They proposed a method for detecting and measuring timing characteristic of micro-expressions. Frame-by-frame classification was done to detect AUs in 8 video cube regions. The \textit{Onset} frame and \textit{Offset} have higher accuracy than the \textit{Apex} frame, which indicates that their proposed descriptor is suitable for recognition rather than classification for a static frame. To measure AU timing characteristics, the change of bin values in the 3D gradient orientation histogram have been used to reflect the changes and motion accelerations of facial movement. They claimed that this time profile could be used to identify the distinction between posed and spontaneous micro-expression. 

Different facial regions having different contributions to micro-expressions as Chen et al. claimed ~\cite{chen2016emotion} and this being largely ignored by previous studies. They proposed to used 3DHOG features with weighted method and used fuzzy classification for micro-expression recognition. They evaluated their method on 36 samples from CASME II, which contains 4 emotions at a rate of 9 samples per emotion. They compared the result with 3DHOG and weighted 3DHOG and perform better than both achieving average accuracy of 86.67\%.                  

\begin{table*}[h]
	\caption{Summary (Part I: 2009 - 2015) of the feature types, classifier and metrics used over the past decade for micro-expression recognition by year and authors.}  
	\label{table:Related-work}
	\centering
	\renewcommand{\arraystretch}{1.2}       
	\resizebox{\textwidth}{!}{
		\begin{tabular}{ | l | l | l | l | l | l | }	\hline
			Year & Authors & Datasets & Feature type & Classifier & Metrics (Best Result) \\ \hline
			2009 & Polikovsky et al. \cite{polikovsky2009facial} & Polikovsky & 3DHOG & K-means & AUs Classification  \\ \hline
			2011 & Pfister et al.\cite{pfister2011recognising} & Earlier version of SMIC & LBP-TOP & SVM, MKL and RF & Accuracy: 71.4\% using MKL \\\hline
			2011 & Pfister et al.\cite{pfister2011differentiating} & SPOS & CLBP-TOP & SVM, MKL and LINEAR & Accuracy: 80\% using MKL\\ \hline
			2013 & \begin{tabular}[c]{@{}l@{}}Polikovsky \\  and Kameda\cite{polikovsky2013facial}\end{tabular} & Polikovsky & 3DHOG & K-means & Recognition of 11 AUs \\ \hline
			2013 & Li et al. \cite{li2013spontaneous} & SMIC(HS, VIS and NIR) & LBP-TOP & SVM & Accuracy: 52.11\% on VIS \\ \hline
			2013 & Song et al. \cite{song2013learning} & SEMAINE corpus & HOG+HOF & SVR & N/A \\ \hline
			2014 & Guo et al. \cite{guo2014micro} & SMIC & LBP-TOP & nearest neighbour & Accuracy: 65.83\% \\ \hline
			2014 & Yan et al. \cite{yan2014casme} & CASME II & LBP-TOP & SVM & Accuracy: 63.41\% \\ \hline
			2014 & Wang et al. \cite{wang2014micro} & CASME and CASME II & TICS & SVM & \begin{tabular}[c]{@{}l@{}}Accuracy: 61.85\% on CASME \\  58.53\% on CASME II\end{tabular}  \\ \hline
			2014 & Le et al. \cite{le2014spontaneous} & CASME II and SMIC & LBP-TOP+STM & AdaBoost & \begin{tabular}[c]{@{}l@{}}Accuracy: 43.78\% on CASME II \\  44.34\% on SMIC \end{tabular}  \\ \hline
			2014 & Lu et al. \cite{lu2014delaunay} &\begin{tabular}[c]{@{}l@{}}SMIC, CASME B \\ and CASME II\end{tabular}   & DTCM & SVM, RF & \begin{tabular}[c]{@{}l@{}} Accuracy:82.86\% on SMIC, 64.95\%\\on CASME and 64.19\% on CASME II\end{tabular}   \\ \hline
			2014 & Liong et al. \cite{liong2014subtle} & CASME II and SMIC & OSW-LBP-TOP & SVM & \begin{tabular}[c]{@{}l@{}} Accuracy:57.54\% on SMIC \\  66.40\% on CASME II \end{tabular}   \\ \hline
			2014 & Wang et al. \cite{wang2014face} & CASME & DTSA & ELM & Accuracy: 46.90\% \\ \hline
			2014 & Davison et al. \cite{davison2014micro} & CASME II & LBP-TOP+GDs & RF, SVM & 92.6 \% when RF used \\ \hline
			2015 & House and Meyer \cite{house2015preprocessing} & SMIC & LGCP-TOP & SVM & Accuracy 48.1\% \\ \hline
			2015 & Wang et al. \cite{wang2015efficient} & CAMSE II and SMIC & LBP-SIP and LBP-MOP & SVM & \begin{tabular}[c]{@{}l@{}} Accuracy:66.8\% on CASME \\  using LBP-MOP \end{tabular}   \\ \hline
			2015 & Wang et al. \cite{wang2015micro} & CASME and CASME II & TICS, CIELuv and CIELab & SVM & \begin{tabular}[c]{@{}l@{}} Accuracy:61.86\% on CASME \\  62.30\% on CASME II \end{tabular}   \\ \hline
			2015 & Le et al. \cite{le2015subtle} & CASME II & DMDSP+LBP-TOP & SVM, LDA & F1-score: 0.52  \\ \hline
			2015 & Huang et al.\cite{huang2015facial} & CASME II and SMIC & STLBP-IP & SVM & \begin{tabular}[c]{@{}l@{}} Accuracy:59.51\% on CASME II \\  57.93\% on SMIC \end{tabular}   \\ \hline
			2015 & Liu et al. \cite{liumain} & \begin{tabular}[c]{@{}l@{}}SMIC, CASME \\ and CASME II\end{tabular} & MDMO & SVM & \begin{tabular}[c]{@{}l@{}}Accuracy:68.86\% on CASME \\ 67.37\% on CASME II and 80\% on SMIC\end{tabular}   \\ \hline
			2015 & Li et al. \cite{li2015reading} & CASME II and SMIC & LBP, HOG and HIGO & LSVM & \begin{tabular}[c]{@{}l@{}}Accuracy:57.49\% on CASME II \\ 53.52\% on SMIC \end{tabular}   \\ \hline
			2015 & Kamarol et al. \cite{kamarol2015spatio} & CASME II & STTM & SVM one-against-one & Accuracy:91.71\% \\ \hline
	\end{tabular}}
\end{table*}

\subsection{Local Binary Pattern-Three Orthogonal Planes (LBP-TOP) and Variations}
Pfister et al.~\cite{pfister2011recognising} proposed a framework for recognising spontaneous facial micro-expressions. LBP-TOP~\cite{zhao2007dynamic} as a spatio-temporal local texture descriptor has been used to extract dynamic features. In classification phase, Support Vector Machine (SVM), Multiple Kernel Learning (MKL) and Random First (RF) have been used. This framework was evaluated on earlier version of SMIC where the data collected from only six participants with 77 sample of micro-expressions. Temporal Interpolation Model (TIM) has been used to increase the number of frames to achieve more statistically stable histograms. The result of SMIC were compared to York Deception Detection Test (YorkDDT) \cite{warren2009detecting} which were recorded in 25 fps and resolution 320$\times$240. Using leave-one-subject-out (LOSO), the method was evaluated on two corpora and down-sampled SMIC to 25 fps. They have two sets to classify between them emotional vs non-emotional and lie vs truthful. The best result achieved on YorkDDT to classify between first set is an accuracy of 76.2\% using MKL and 10 frames. For the second set, the best result is 71.5\% using MKL 10 frames and same result using SVM. For SMIC they classify between negative and positive, the best result is 71.4\% using MKL 10 frames and 64.9\% using MKL and 15 frames for down-sampled SMIC.   

Pfister et al.~\cite{pfister2011differentiating} then proposed a method to differentiate between spontaneous and posed facial expressions (Spontaneous Vs Posed (SVP)). They extended Complete Local Binary Patterns (CLBP) which was proposed by Guo et al. \cite{guo2010completed} to work with dynamic texture descriptor and called it CLBP from Three Orthogonal Planes (CLBP-TOP). They evaluated their proposed method by leave-one-subject-out on a corpus developed by them Spontaneous vs POSed (SPOS). This SPOS provides spontaneous and posed expression for the same subject in the session. It contains 7 subjects with 84 posed and 147 spontaneous expressions. Two cameras have been used to record the corpus, one recorded data from visual (VIS) and the other from near-infrared channel (NIR). Both of cameras used 640$\times$480 resolution and 25 fps. SVM, LINEAR classifier (LIN), Multiple Kernel Learning (MKL) and fusion of SVM, LIN and Random Forest through majority voting (FUS) have been used as classifiers. They showed that CLBP-TOP overcome LBP-TOP with an accuracy of 78.2\%, 72\% and 80\% on NIR, VIS and combination, respectively.

Li et al. \cite{li2013spontaneous} run two experiments on SMIC database for analysing micro-expressions. The first experiment was to detect micro-expressions occurring and the other was to then recognise the type of micro-expression. The detection stage was employed to distinguish a micro-expression and a normal facial expression. On the other hand, recognition discriminated three classes of micro-expression (positive, negative and surprise). A normalisation was done to all faces, followed by a registration to a face model using 68 feature points from an Active Shape Model \cite{cootes1995active}. Then the faces were cropped according to the eye positions that has been detected using Haar eye detector \cite{niu20062d}. LBP-TOP was used for feature extraction from cropped face sequences. 

In the VIS and NIR dataset which has a limited number of frames, some problems may arise when applying LBP-TOP. To avoid these problems, TIM was used to allow up-sampling and down-sampling of the number of frames. SVM was used as a classifier and leave-one-subject-out cross validation \cite{wu2011machine} was used to compute the performance of the two experiments, which were run on three datasets (HS, VIS and NIR). The best accuracy for detection of micro-expressions was 65.55\% when evaluating the method on the HS dataset and the X, Y and T parameters were equal to 5, 5 and 1 respectively for LBP-TOP. For micro-expressions recognition, the best accuracy is equal to 52.11\% on VIS dataset with X, Y and T having the same value as previous. Avoiding the problem that may arise because the limitation regarding the number of frames by using TIM is considered a strength for this algorithm. However, there is a limitation in using a limited number of recognition classes, since some emotion cannot be judged under ambiguous conditions if more than one expression reported by the participant.

Guo et al. \cite{guo2014micro} used LBP-TOP features in their micro-expression recognition experiment. To classify these features, they used the nearest neighbour method to compare the distance between unknown samples with entire known samples. Euclidean distance has been used as distance measurement. This method was evaluated on SMIC database. In evaluation, firstly they used Leave-One-Subject-Out (LOSO) and Leave-One-Expression-Out (LOEO) and achieved a recognition accuracy of 53.72\% and 65.83\% for LOSO and LOEO respectively. In addition, they have conducted experiments for different values of LBP-TOP parameters ($R_{X},R_{Y},R_{T},P_{X},P_{Y},P_{T}$) which refer to the radii in axis X, Y and T, and the number of neighbourhood points in the XY, XT and YT planes respectively. The best result was achieved when they set the value to (1,1,2,8,8,8) for parameters. A different distribution of training set and testing set also have been tested and the best result of 63\% was achieved when portion of training and testing data with a 5:1 split.

Yan et al. \cite{yan2014casme} carried out a micro-expression recognition experiments on clips from the CASME II dataset, developed by the same authors. LBP-TOP was used in this experiment to extract the features. SVM was employed as the classifier. With radii varying from 1 to 4 for X and Y, and from 2 to 4 for T (they do not consider T=1 due to little change between two neighbouring frames on a sample rate of 200 fps), and SVM was used as the classifier which classify between five main categories of emotions provided in this experiment (happiness, disgust, surprise, repression and others). The best performance is 63.41\% shown when the radii are 1, 1 and 4 for XY, YT and XT planes respectively. Developing high quality datasets with higher temporal (200 fps) and spatial resolution (about 280$\times$340 pixels on facial area), and classify 5 categories of expression with performance 63.41\% are the advantages of this method, however they use same method which used for classifying ordinary facial expressions which may not work well for micro-expressions.

Liong et al. \cite{liong2014subtle} proposed Optical Strain Weighted LBP-TOP (OSW-LBP-TOP) method which used optical strain features for micro-expression recognition. They evaluated this feature on CASME II and SMIC. They used SVM as classifier and test different kernel. Their method outperformed the two baseline methods \cite{yan2014casme,li2013spontaneous} when evaluated on two datasets and achieved accuracy of 57.54\% on SMIC when using poly kernel and 66.40\% on CASME II when RBF kernel was used.

Davison et al. \cite{davison2014micro} developed a method to differentiate between micro-movements (MFMs) and neutral expression. This method has been evaluated on CASME II database. LBP-TOP and Gaussian Derivatives (GDs) features are obtained. RF and SVM used as classifiers. Normalization has been done before extract the features to make sure that all faces are in the same position. The images have been divided into 9x8 blocks with no overlapping. Local features obtained for each block after being processed separately using GDs. These local features concatenated into the overall global feature description. LBP-TOP has been calculated for each block through all three planes X, Y and T. In the classification phase data has been separated into testing and training. 100-fold cross-validation was used for testing. The best accuracy achieved is 92.60\% when RF has been used and separate testing and training data into 50\% and a combination of LBP-TOP and GDs features were used.

House and Meyer \cite{house2015preprocessing} implemented a method for micro-expression recognition and detection. They used LBP-TOP and local gray code patterns on three orthogonal planes (LGCP-TOP) as features descriptors. SVM has been used as classifier and SMIC database used to evaluate the method. LGCP-TOP is modified version of LGCP \cite{islam2013local} that originally worked for facial expressions and re-worked for analysing the dynamic texture of micro-expressions. They did not overcome the result of LBP-TOP from \cite{li2013spontaneous} and they returned this to the feature vectors of LGCP-TOP, which is too large to be classified without over-fitting. They claimed that LGCP-TOP had advantage over LBP-TOP in computational time of the feature descriptor.

Wang et al. \cite{wang2015efficient} inspired two feature descriptors for micro-expressions recognition from the concept of LBP-TOP, LBP-Six Intersection Points (SIP) and LBP-Three Mean Orthogonal Planes (MOP). LBP-SIP is an extension of LBP-TOP and more compact form. This compaction is based on the duplication in computing neighbour points through three planes. Therefore, they only considered the 6 unique points on intersection lines of three orthogonal planes. They claimed that these 6 points carry sufficient information to describe dynamic textures. Vector dimensions in LBP-SIP is 20, in contrast LBP-TOP produce 48 dimensions.

The basic idea of LBP-MOP is to compute features of mean planes rather than all frames in the video. Those two descriptors were evaluated on CASME II and SMIC databases use baseline settings for both datasets \cite{yan2014casme, li2013spontaneous}. Leave-one-video-out (LOVO) and Leave-one-subject-out (LOSO) cross-validation configurations have been tested on two datasets with different popular kernels for SVM. Also a Wiener filter has been applied for image smoothing to remove noise. LBP-MOP achieved best result (66.8\%) on CASME II with linear kernel for SVM using LOVO cross validation and Wiener filter applied in preprocessing step. On SMIC the two methods did not achieve better results than the original LBP-TOP, which achieved 66.46\% with Wiener filter and RBF kernel for SVM using LOVO cross validation.    

Wang et al. \cite{wang2014micro} proposed a novel color space model for micro-expressions recognition using dynamic textures on Tensor Independent Color Space (TICS) in which the color components are as independent as possible. They claimed it will enhance the performance of micro-expression recognition. It differs from other literature \cite{yan2014casme} \cite{li2013spontaneous} in getting LBP-TOP features from color as a fourth-order tensor in addition to width, height and time. These experiments were conducted on two micro-expression databases, CASME and CASME II. SVM has been used as classifier. The results show that the performance in TICS is better than that in RGB or grayscale, where the best result achieved on CASME class B is 61.85\% and 58.53\% on CASME II. Although the accuracy is lower than other state of the art in same area \cite{yan2013casme, yan2014casme} but reveals that TICS may provide useful information more than RGB and grayscale.

In addition to TICS, Wang et al. \cite{wang2015micro} further show that CIELab and CIELuv are also could be helpful in recognising micro-expressions. They achieved 61.86\% accuracy on CASME class B using TICS, CIELuv and CIELab with different parameters for LBP-TOP. An accuracy of 62.30\% was achieved on CASME II using TICS and CIELuv with different parameters for LBP-TOP.  

Le et al. \cite{le2015subtle} proposed a preprocessing step that may enhance recognition rate for micro-expressions. Due to the redundant frames without significant motion which generated when recording with high-speed camera which have high fps, they proposed to use Sparsity-Promoting Dynamic Mode Decomposition (DMDSP) \cite{jovanovic2014sparsity} to analyse and eliminate this redundancy. They used LBP-TOP to extract features, with SVM and Linear Discriminant Analysis (LDA) \cite{chien2005linear} as classifiers. This method was evaluated on CASME II. F1-score, recall and precision have been used to measure the performance. The percentages of reserved frame using DMDSP were varied between 45\% and 100\% of original frame length. The performance increased while the percentages of reserved frames decreased. The best performance was achieved when 45\% of frames were reserved with F1-score, precision and recall equal to 0.52, 0.48 and 0.56 respectively when SVM was used, and 0.47, 0.42 and 0.53 when LDA was used. The performance was compared to the benchmark of CASME II \cite{yan2014casme} and outperformed the benchmark.

Le et al. \cite{le2014spontaneous} defined three difficulties that faced Micro-Expression recognition systems: difficulty of being able to differentiate between two micro-expressions for one subject, namely inter-class similarity, dissimilarity of the same micro-expression between two subjects due to the different facial morphology and behaviour, the uneven distribution of each classes and subjects.
They aimed to resolved two latter problems by using facial registration, cropping and interpolation as preprocessing to remove morphological differences. They have proposed variant of AdaBoost to deal with imbalanced characteristics of micro-expressions. The experiments were evaluated on CASME II and SMIC. In addition, TIM has been used to avoid the biases that can be caused by the different frame lengths. For feature extraction LBP-TOP was used and a Selective Transfer Machine (STM) has been used to avoid imbalances which came from the mismatch between distributions of training and testing samples that caused by leave-one-subject-out (LOSO) cross validation to evaluate the datasets. The best result was achieved on CASME II (43.78\% recognition rate) when STM used with AdaBoost and fixed frame length of 15 frames, for SMIC, 10 frames give the best result (44.34\% recognition rate).

More recently, Talukder et al. \cite{talukder2016intelligent} used LBP-TOP as features extraction and SVM as classifier after magnified the motion to enhance the low intensity of micro-expression. They conducted their method on the SMIC dataset. They claimed that there is improvement on the recognition result due to the motion magnification applied with average recognition rate up to 62\% on SMIC-NIR.

Unlike other studies Duan et al. \cite{duan2016recognizing} extracted LBP-TOP from the eye region, not from the whole face. They tested this method on CASME II. They used more than 20 classifiers to train the features. Their method performed better than other methods when classifying happy and disgust expressions.  

Huang et al.~\cite{huang2015facial} proposed Spatio-Temporal Local Binary Pattern with Integral Projection (STLBP-IP). They used integral projection to boost the capability of LBP-TOP with experiments conducted on the CASME II and SMIC datasets using SVM as a classifier. When they tested this method on CASME II, it was been compared with several methods from different studies and was used different parameters for LBP-TOP and different kernel for SVM, and also compared with LBP-SIP \cite{wang2015efficient} and LOCP-TOP \cite{spatiotemporal2011local} that achieved a promising performance over these methods with an accuracy rate of 59.51\%. When they evaluated their method on SMIC they compared it with \cite{dollar2005behavior,jain2011facial,zhao2007dynamic,li2013spontaneous} and achieved 57.93\%.

Huang et al. \cite{huang2016spontaneous2} proposed facial micro-expression recognition method using discriminative spatio-temporal local binary pattern with an improved integral projection. They proposed this method to preserve the shape attribute of
micro-expressions. They claimed that extracting features from the global face region lead to ignoring the discriminative information between different classes. They conducted this method on three publicly available datasets: SMIC, CASME and CASME II. They compared this new method with their previous study \cite{huang2015facial} and demonstrated better results across three datasets with accuracy rate up to 64.33\% on CASME, 64.78\% on CASME II and 63.41\% on SMIC.

Wang et al. \cite{wang2016effective} used LBP-TOP features to recognise micro-expressions after pre-processed the CASME II dataset with Eulerian Video Magnification (EVM). SVM and k-nearest neighbour (KNN) have been used as classifiers to classify between 5 motions from CASME II dataset. They used leave-one-subject validation with comparison with baseline \cite{yan2014casme} and other methods \cite{wang2015efficient,wang2014lbp,park2015subtle,liong2014subtle}. Their proposed method achieved accuracy of up to 75.30\%. 

Zhang et al. \cite{zhang2017micro} combined local LBP-TOP and local Optical Flow (OF) features after extracted them from local regions of face based on AUs and conducted it on CASME II. They claimed that different local features can perform better than single global features. They compare between different classifiers with different parameters (KNN, SVM and RF), also a comparison between global features and local features has been conducted to prove their hypothesis. Accuracy up to 62.50\% has been achieved when they combined two local features with RF classifier. 

To solve the cross-database ME recognition problem Zong et al. \cite{zong2017learning} proposed a method to regenerate the target sample in the process of recognition to have the similar feature distributions as source sample, they called their method Target Sample Re-Generator (TSRG). They evaluated this method on CASME II and three types of SMIC, therefore six experiments have been conducted where the databases served as source and target. Uniform LBP-TOP have been used as features extractor and UAR and WAR used as performance measurement. Comparing to some state-of-the-art method TSRG overcome them in seven experiments in both weighted average recall (WAR) and unweighted average recall (UAR) of 12 in total. They improve their work in \cite{zong2018domain} and proposed a frame work called it Domain Regeneration (DR) the difference is the generating from both source and target for more similar feature distributions. And they used here three domains to regenerating samples DR-face space for target (DRFS-T), DR-face space for sample (DRFS-S) and DR-line space (DRLS).

By combining heuristic and automatic approaches Liong et al. \cite{liong2018hybrid} introduced a method to recognize micro-expression by selecting facial regions statically based on AUs frequency occurring(ROI-selective). They used a hybrid features (Optical Strain Flow (OSF) and block-based LBP-TOP).  They tested their method on CASME II and SMIC using SVM as classifier with LOSOCV and LOVOCV to validate the effectiveness. The results have been reported using more than one measurements including accuracy and F-measure and compared with baseline of OSF and LBP-TOP. the method overcome the baseline of two features in all measurement and with both validations, in term of F-measure the best result was 0.51 and 0.31 on SMIC and CASME II respectively.

Zong et al. \cite{zong2018learning} argued that extracting features of fixed-sized facial blocks for micro-expression recognition is not suitable technique. This is due to the fact that it may ignore some information about the AU if it is small or may get overlapping if it is large, leading to the extraction of confusing information. To solve the mentioned problem, they proposed hierarchical division scheme which is dividing face into regions with different densities and different size. They also proposed a learning model called it kernelized group sparse learning (KGSL). More than one feature types have extracted from those hierarchical divisions such as LBP-TOP, LBP-SIP and STLBP-IP. Evaluating of hierarchical division and KGSL have been done on CASME II and SMIC using LOSOCV. The best result achieved on CASME II, when using Hierarchical STLBP-IP + KGSL and it was 63.97\% and 0.6125 in term of accuracy and F1-score respectively.    

\begin{table*}[h]
	\caption{Summary (Part II: 2016 - 2018) of the feature type, classifier and metrics used over the past decade for micro-expression recognition by year and authors.}  
	\label{table:Related-workII}
	\centering
	\renewcommand{\arraystretch}{1.2}       
	\resizebox{\textwidth}{!}{
		\begin{tabular}{ | l | l | l | l | l | l | }	\hline
			Year & Authors & Datasets & Feature type & Classifier & Metrics (Best Result) \\ \hline
			2016 & Chen et al.\cite{chen2016emotion} & CASME II(36 samples) & 3DHOG & Fuzzy &Accuracy: 86.67\%. \\ \hline
			2016 & Talukder et al. \cite{talukder2016intelligent} & SMIC & LBP-TOP & SVM &Accuracy: 62\% on SMIC-NIR \\ \hline
			2016 & Duan et al. \cite{duan2016recognizing} & CASME II & LBP-TOP from eye region & 26 classifiers & Perform better on happy and disgust \\ \hline
			2016 & Huang et al. \cite{huang2016spontaneous2} & \begin{tabular}[c]{@{}l@{}}SMIC, CASME \\ and CASME II\end{tabular} & improved of STLBP-IP & SVM &\begin{tabular}[c]{@{}l@{}}Accuracy:64.33\% on CASME \\ 64.78\% on CASME II and 63.41\% on SMIC\end{tabular}  \\ \hline
			2016 & Wang et al. \cite{wang2016effective} & CASME II & LBP-TOP &  SVM and KNN & Accuracy: 75.30\% \\ \hline
			2016 & Zhang et al. \cite{zhang2016micro} & CASME II & gabor filter+ PCA and LDA & SVM & Good performance on static image \\ \hline
			2016 & Huang et al. \cite{huang2016spontaneous} & \begin{tabular}[c]{@{}l@{}}SMIC, CASME \\ and CASME II\end{tabular} & STCLQP & Codebook & \begin{tabular}[c]{@{}l@{}}Accuracy:64.02\% on SMIC\\  57.31\% CASME and 58.39\% CASME II \end{tabular}\\ \hline
			2016 & Ben et al. \cite{ben2016gait} & CASME & MMPTR & Euclidean distance & Accuracy: 80.2\% \\ \hline
			2016 & Liong et al. \cite{liong2016less} & SMIC and CASME II & Bi-WOOF & SVM & F1-score:0.61 on CASME II, 0.62 on SMIC-HS \\ \hline
			2016 & Liong et al. \cite{liongautomatic} & SMIC and CASME II & Bi-WOOF & SVM & \begin{tabular}[c]{@{}l@{}} F1-score:0.59 on CASME II \\ Accuracy:53.52 on SMIC-VIS \end{tabular}  \\ \hline
			2016 & Liong et al. \cite{liong2016spontaneous} & CASME II and SMIC & Optical Strain & SVM & \begin{tabular}[c]{@{}l@{}}Accuracy:63.41\% on CSME II \\ 52.44\% on SMIC\end{tabular}\\ \hline
			2016 & Oh et al. \cite{oh2016intrinsic} & CASME II and SMIC & I2D & SVM & F1-score: 0.41 and 0.44 on CASME II and SMIC \\ \hline
			2016 & Wang et al. \cite{wang2016sparse} & CASME and CASME II & STCCA & Nearest Neighbor, SVM & \begin{tabular}[c]{@{}l@{}}Mean recognition accuracy : 41.20\% on CASME\\ 38.39 on CASME II \end{tabular}\\ \hline
			2016 & Zheng et al. \cite{zheng2016relaxed} & CASME and CASME II & LBP-TOP, HOOF & RK-SVD & \begin{tabular}[c]{@{}l@{}}Accuracy:69.04\% on CASME\\ 63.25\% on CASME II \end{tabular} \\ \hline
			2016 & Kim et al. \cite{kim2016micro}  & CASME II & CNN & LSTM & Accuracy: 60.98\% \\ \hline
			2017 & Zhang et al. \cite{zhang2017micro} & CASME II & LBP-TOP,Optical Flow & KNN, SVM and RF & Accuracy: 62.50\% \\ \hline
			2017 & Zheng \cite{zheng2017micro} & \begin{tabular}[c]{@{}l@{}}SMIC, CASME \\ and CASME II\end{tabular} & 2DGSR & SRC & \begin{tabular}[c]{@{}l@{}}Accuracy:71.19\% and 64.88\% \\ on CASME and CASME II\end{tabular}  \\ \hline
			2017 & Ben et al. \cite{ben2017learning} & CASME II & HWP-TOP & SSVM & \begin{tabular}[c]{@{}l@{}}Recognition rate of 0.868\end{tabular}  \\ \hline
			2017 & Zong et al. \cite{zong2017learning} & CASME II and SMIC & LBP-TOP & TSRG & \begin{tabular}[c]{@{}l@{}}UAR 60.15\end{tabular}  \\ \hline
			2017 & Happy and Routray \cite{happy2017fuzzy} &\begin{tabular}[c]{@{}l@{}} CASME II, \\ CASME and SMIC \end{tabular} & FHOFO & SVM, KNN and LDA & \begin{tabular}[c]{@{}l@{}}F1-score was 0.5489, 0.5248 and 0.5243 \\ CASME, CASME II and SMIC \end{tabular}  \\ \hline
			2017 & Hao et al.  \cite{hao2017deep}  & JAFFE  & WLD and DBN & DBN & \begin{tabular}[c]{@{}l@{}}Recognition rate: 92.66 \end{tabular}  \\ \hline
			2017 & Peng et al. \cite{peng2017dual} & CASMEI/II  & OF & DTSCNN & \begin{tabular}[c]{@{}l@{}}Accuracy up to 66.67\% \end{tabular}  \\ \hline
			2018 & Liong et al. \cite{liong2018hybrid} & CASME II and SMIC  & OSF and LBP-TOP & SVM & \begin{tabular}[c]{@{}l@{}}F-measure: 0.51 and 0.31  SMIC and CASME II \end{tabular}  \\ \hline
			2018 & Zhu et al. \cite{zhu2018coupled}  & CASME II & LBP-TOP and OF & SVM & \begin{tabular}[c]{@{}l@{}}accuracy of 53.3\% \end{tabular}  \\ \hline
			2018 & Zong et al. \cite{zong2018learning} & CASME II and SMIC& LBP-TOP, LBP-SIP and STLBP-IP & KGSL & \begin{tabular}[c]{@{}l@{}}F1: 0.6125 on CASME II  \end{tabular}  \\ \hline
	\end{tabular}}
\end{table*}

\subsection{Histogram of Oriented Optical Flow (HOOF)} 
Liu et al. \cite{liumain} proposed Main Directional Mean Optical-flow (MDMO) as features for recognition micro-expression. Their MDMO consist of Regions of Interest (ROIs) based partially on AUs. One of the significant advantages of MDMO is the small features dimension, where the features vector length equal to 72 which is 2 features extracted from each region of 36 ROIs.
Aligned all frames to the first frame has been applied to reduce the noise result from head movements. SVM classifier has been adopted for recognition. SMIC, CASME and CASME II datasets were used to evaluate their method. The result compared to the benchmark which used LBP-TOP and histogram of oriented optical flow (HOOF) features and achieve better result compare to benchmark which 68.86\%, 67.37\% and 80\% on CASME, CASME II and SMIC respectively. 

Song et al. \cite{song2013learning} used a Harris3D detector with combination of HOG and the Histograms of Oriented Optical Flow (HOOF) features, and used codebook to encode features in a sparse manner of micro-expressions. To predict expression they used Support Vector Regression (SVR) \cite{smola2004tutorial}. They evaluated this method on a subset of the SEMAINE corpus \cite{mckeown2012semaine} dataset.

Happy and Routray \cite{happy2017fuzzy} they claimed that the changes on the face during a micro-expression is temporal changes more than spatial. Based on this claim they proposed temporal features descriptor called Fuzzy Histogram of Optical Flow Orientation (FHOFO) and it’s an extension of HOOF. They evaluated their method on CASME, CASME II and SMIC. The best result was achieved in term of F1 score was 0.5489, 0.5248 and 0.5243 on the mentioned datasets respectively. In \cite{happy2018recognizing} They used Pair-wise feature proximity (PWFP) as features selection to improve the result in the previous study which has been slightly improved.

To enhance micro-expression recognition Zhu et al. \cite{zhu2018coupled} transfer learning from speech to micro-expression and call their method coupled source domain targetized with updating tag vectors. LBP-TOP and OF have been used as features extractor with different vector dimension. They used SVM as classifier and evaluated their method on CASME II. The best accuracy of 53.3\% achieved by OF with dictionary dimensions at 50.

\subsection{Deep Learning Approaches}
Over the past few years, deep learning approaches, such as convolutional neural networks (CNNs), have grown rapidly with a growing number of successful applications \cite{lecun2015deep,deng2014deep}. A core feature of CNNs is the network architecture that produces the features to represent the input data. Popular architectures include LeNet \cite{lecun1998gradient}, GoogLeNet \cite{szegedy2015going} and AlexNet \cite{krizhevsky2012imagenet}. Many deep learning approaches focus on static images for classification, object detection or segmentation. Spatio-temporal based analysis methods using 3D CNNs are emerging with new applications, primarily on action recognition \cite{ji20133d,karpathy2014large,yue2015beyond,tran2015learning}.

As the datasets associated with these new methods are very large in number, for example the Sports-1M Dataset by Karpathy et al. \cite{karpathy2014large}, gaining discriminative data for 3D CNNs is a much easier task than collecting spontaneously induced micro-expressions. Therefore, there are very few approaches to detecting and recognising subtle motion using deep learning.

One of the first to use CNNs in micro-expressions analysis is by Kim et al. \cite{kim2016micro}. They proposed a new feature representation for micro-expressions where the spatial information at different temporal states (i.e. onset, apex and offset) are encoded using a CNN. This method used the extracted features attempted to help discriminate micro-expression classes when the model is passed to the long short-term memory (LSTM) recurrent neural network, where the temporal characteristics of the data are analysed. The overall achieved accuracy when comparing with the state-of-the-art was 60.98\%, which is still relatively similar to many micro-expression recognition systems that only use accuracy for the evaluation metric. Further, the method only evaluated on single dataset, i.e. CASME II \cite{yan2014casme} dataset and does not consider more modern micro-expression datasets such as SAMM \cite{davison2018samm} and CAS(ME)$^2$ \cite{qu2017cas}.

In 2017, Peng et al. \cite{peng2017dual} proposed a new method named Dual Temporal Scale Convolutional Neural Network (DTSCNN). Due to the data deficiency in available datasets, they designed a shallower neural network for micro-expression recognition with only 4 layers for both convolutional and pooling layers. As stated in its name, DTSCNN is a two streams network. The network has been fed with the optical-flow sequences. CASMEI/II datasets were used in the experiment and have been merged by the authors using selected data from both datasets, CASME I/II have been categorized into 4 classes: Negative, Others, Positive and Surprise. They achieved the best accuracy of 66.67\%. 

Hao and Tian \cite{hao2017deep} used deep belief network (DBN) as the second stage features extractor to extract more global feature with less computation cost. DBN classification has been done by pre-training and fine-tuning in DBN. This was fused with the first stage local features was Weber Local Descriptor (WLD).  However, their method only evaluated on a non-spontaneous dataset JAFFE \cite{lyons1998japanese}, which was dated and difficult to compare with current literature.

The review reflects the existing CNN-based methods faced similar problem in terms of data. Overall, micro-expression recognition using deep learning is still in its infancy due to a lack of available dataset. A large amount of data is crucial when training CNNs like many machine learning approaches. Micro-expressions are very complex and cannot easily be categorised into distinct classes as many approaches attempt to do~\cite{pfister2011recognising, yan2013how,yap2014facial}. Using 3D CNN features to understand the subtle movement would be a better approach to generalise the problem of discriminating a micro-expression on the face.

\subsection{Other Feature Extraction Methods}
Lu et al. \cite{lu2014delaunay} proposed Delaunay-Based Temporal Coding Model (DTCM) for Micro-Expression Recognition. Active Appearance Model (AAM) used to define facial feature points (68 points). Delaunay triangulation has been implemented based on the feature points. This process divides the facial area into number of sub-regions with triangle shape. Normalisation has been done based on standard face (neutral), this remove personal appearance difference irrelative. They used local temporal variations (LTVs) to code the features space, where the difference between mean of grayscale values of subregion and sub-region in neighbour frame were computed. Delaunay triangulation generates a large number of subregions which leads to large number of local features. To overcome this problem, they selected just subregions related to micro-expression, this selection based on standard deviation analysis. finally, the code sequences of all subregions concatenated into one feature vector. RF~\cite{breiman2001random} and SVM have been used as classifiers. This method evaluated on SMIC, CASME class B and CASME II. They achieved better result than state of the art, with 82.86\%, 64.95\% and 64.19\% on SMIC, CASME class B and CASME II respectively.

Zhang et al. \cite{zhang2016micro} developed micro-expression recognition system or visual platform as they claimed that there has not been much work done in designing these kind of systems. Their system includes two main parts: feature extraction and dimensional reduction, they used a gabor filter for feature extraction and principal components analysis (PCA) and LDA for dimension reduction. For classification stage, SVM has been used. To evaluate their system, CASME II and real-time videos were used. They claimed that the system have a good performance on static images counter to real-time videos. Gabor filter also been used by Wu et al. \cite{wu2011machine} but they evaluated the performance on Cohn and Kanade’s dataset (CK) \cite{kanade2000comprehensive}, which was developed for facial expression analysis.          
 
Li et al. \cite{li2015reading} evaluated the performance of three feature types (LBP, HOG and histograms of image gradient orientation (HIGO)) on two publicly available datasets (CASME II and SMIC). They extracted these three features from different planes. LSVM was employed as the classifier using LOSO validation. On CASME II, the best accuracy was 57.49\%. This is achieved when they extracted HOG from both 3 orthogonal planes (HOG-TOP) and XT, YT planes (HOG-XYOT). On the other hand, three versions of SMIC were tested - VIS, NIR, HS and sub of HS, with the last one achieved the best accuracy and when features were extracted using HIGO-TOP and HOG-TOP. In addition, an effect of the interpolation length was tested with different frame lengths from 10 to 80 with fixed incremental steps of 10 frames. The best performance was achieved with an interpolation to 10 frames and it was 53.52\%, 45.12\% and 38.02\% on SMIC-VIS, SMIC-HS and SMIC-NIR respectively.

Huang et al. \cite{huang2016spontaneous} outlined two problems of LBP-TOP. The first problem is LBP-TOP does not consider useful information, the second problem is the classical pattern used by LBP-TOP may not be good for describing local structure. To avoid those two problems, they proposed Spatio-Temporal Completed Local Quantization Patterns (STCLQP), which extracted sign, magnitude and orientation. In addition, a codebook were developed for each component in both appearance and temporal domains.Their method was evaluated on SMIC, CASME and CASME II with accuracy of 64.02\%, 57.31\% and 58.39\%, respectively.

Spatio-Temporal Texture Map (STTM) was developed by Kamarol et al. \cite{kamarol2015spatio}. STTM used a modified version of Harris corner function \cite{harris1988combined} to extract the micro-expression features. This method evaluated on CASME II, and compared with other features (Volume Local Binary Pattern (VLBP) and LBP-TOP). They used SVM with one-against-one classification between four classes. In terms of accuracy, the average recognition rate of STTM performed slightly better than the other features which is reached to 91.71\% in contrast LBP-TOP achieved 91.48\%. On the other hand, in terms of computation time, there is a large difference between STTM and other features, where STTM process one frame in 1.53 seconds in contrast to 2.57 and 2.70 seconds for VLBP and LBP-TOP respectively. 

Wang et al. \cite{wang2014face} introduced a micro-expression algorithm called discriminant tensor subspace analysis (DTSA). This method was evaluated on the CASME dataset. Extreme learning machine (ELM) was used as classifier. They have tested the method with various optimal dimensionality and different sets of training and testing. The best accuracy, 46.90\%, was achieved when dimensionality was set to 40$\times$40$\times$40 and the training sample is 15.   

Maximum margin projection with tensor representation (MMPTR) is a micro-expression recognition algorithm contributed by Ben et al. \cite{ben2016gait}. They tested their algorithm on CASME. The best average recognition rate, which is 80.2\% was achieved on tensor size of 64$\times$64$\times$64 and training sample was same as \cite{wang2014face} with 15 samples. 

Liong et al. \cite{liong2016less} questioned whether all frames of micro-expressions need to be processed for effective analysis. They used only the apex and the onset frame for experiments to test this theory. The frames were extracted using their proposed Bi-Weighted Oriented Optical Flow (Bi-WOOF). These features were then evaluated on CASME II and the three formats of SMIC. The best performance achieved on CASME II and SMIC-HS in terms of F1-score was 0.61 and 0.62 respectively. Bi-WOOF also used by Liong et al. \cite{liongautomatic} to extract features from just the apex frame after proposing a method using an eye mask to spot it. This method was evaluated on CASME II and SMIC and achieved 0.59 on CASME II in terms of F1-score. 

Liong et al. \cite{liong2016spontaneous} proposed two sets of features: optical strain and optical strain weighted. These two features constructed by utilising facial optical strain magnitudes. They performed the features on the CASME II and SMIC and they overcame the baseline of two datasets \cite{li2013spontaneous,yan2014casme} with recognition rate reach to 52.44\% on SMIC and 63.16\% on CASME II.            

Oh et al. \cite{oh2016intrinsic} claimed that there is changes on facial contour which are located in different part of face are crucial for the recognition micro-expressions. According to that they proposed a feature extraction method to represent these changes called Intrinsic Two-Dimensional local structuresm (I2D). This method was evaluated on the CASME II and SMIC dataset.The result was better than two state of the art \cite{li2013spontaneous,yan2014casme} with the best F1-score of 0.41 and 0.44 on CASME II and SMIC respectively.

Sparse Tensor Canonical Correlation Analysis (STCCA) was proposed by Wang et al. \cite{wang2016sparse} to improve the recognition rate of micro-expressions. They conducted the experiment on CASME and CASME II. They proved that their method can perform better than 3D-Canonical Correlation Analysis and three-order Discriminant Tensor Subspace Analysis. In addition to that they proved that Multi-linear Principal Component Analysis is not suitable for micro-expression recognition.

Zheng et al. \cite{zheng2016relaxed} proposed a a relaxed K-SVD classifier (RK-SVD) and tested it on LBP-TOP and HOOF features to be used for micro-expression recognition. They evaluated this proposed classifier on CASME and CASME II, and compared it with different classifiers such as SVM, MKL and RF. The results was better than other classifiers for both features and on two datasets \cite{yan2013casme,yan2014casme} with best accuracy of 69.04\% and 60.82\% for LBP-TOP and HOOF respectively on CASME, and on CASME II the accuracy was 63.25\% 58.64\% for the same features respectively.

Zheng~\cite{zheng2017micro} proposed a method for micro-expression recognition named 2D Gabor filter and Sparse Representation (2DGSR). They evaluated their method on three publicly available datasets (SMIC, CASME and CASME II) and compared it with other popular methods (LBP-TOP, HOOF-whole and HOOF-ROIs). For classification Sparse Representations Classifier (SRC) has been used with LOSO cross validation. In terms of accuracy they achieved a result up to 71.19\% and 64.88\% on CASME and CASME II respectively.

Ben et al. \cite{ben2017learning} proposed local binary feature descriptor called hot wheel patterns from three orthogonal planes (HWP-TOP) which has been inspired by dual-cross patterns from three orthogonal planes (DCP-TOP) with some rotations. They used smooth SVM (SSVM) as a classifier. They evaluated their descriptor on 61 samples from CASME II with three classes(except fear and sadness) and achieved recognition rate of 0.868. They try to solve the problem of micro-expression limited samples by leverage labeled macro-expression and shared feature between macro and micro expression, however this may be not so accurate due the difference between macro and micro characteristic.

After extracting features using distance estimation between points which have been predicted using ASM Jain et al \cite{jain2018random} using Random Walk-based (RW) to learn the features before providing it to Artificial Neural Network (ANN) classifier. RW reduces the dimensionality of the feature and this minimize the complexity of computation. They evaluated their method on CASME and SMIC and provide the result in term of AUC, which is up to 0.8812 and 0.9456 on SMIC and CASME respectively.  

As shown in this section many methods and feature descriptors have been used for micro-expression recognition, summarization of these methods shown in Table \ref{table:Related-work}. These methods have been evaluated on different datasets which have varying on properties such as frame rates and resolution. In this paper, we contribute to the research by addressing the challenge of how different features react on spatial temporal settings, particularly focusing on resolutions, which has not done in previous research.

\section{Performance Metrics and Validation Techniques}
The spotting accuracy of humans peaks around 40\% \cite{frank2009behavior}. Analysis using computer algorithms incorporating machine learning and computer vision can only be evaluated fairly with a standardised metrics. This section elaborates the metrics used in the literature. Drawing from detailed review in Section 3, we summarised and explain the evaluation metrics. 

\subsection{Metrics}
The metrics for micro-expressions analysis are commonly used for binary classification purposes, and so is adequate for quantifying \textit{True Positive (TP)}, \textit{False Positive (FP)}, \textit{True Negative (TN)} and \textit{False Negative (FN)} detections. More detailed information on these measures can be found in~\cite{baldi2000assessing}. The earlier work, as illustrated in Table \ref{table:Related-work}, the majority of the results in micro-expressions analysis are based on \textit{Accuracy}. as defined in equation \ref{eq:Acc}.  
\begin{equation}
Accuracy = \frac{TP+TN}{TP+FP+TN+FN}
\label{eq:Acc}
\end{equation}
In the later stage, as illustrated in Table \ref{table:Related-workII}, the measurement of performance were reported in \textit{F1-Score (or F-Measure)}. Other metrics such as \textit{Recall}, \textit{Precision},  and \textit{Matthews Correlation Coefficient} (\textit{MCC}) are also gradually used to report the results.
By using the \textit{Precision} measure of exactness, and determines a fraction of relevant responses from results. Recall, or sensitivity, is a fraction of the results that are relevant to the experiment and that are successfully retrieved.
\begin{equation}
Precision = \frac{TP}{TP+FP}
\label{eq:Pre}
\end{equation}
\begin{equation}
Recall = \frac{TP}{TP+FN}
\label{eq:Recall}
\end{equation}

It is unlikely to use these measures on their own as both these measure are commonly used together to form an understanding of the relevance of the results returned from experimental classification.
The F-Measure is useful in determining the harmonic mean between the \textit{Precision} and \textit{Recall} and is used in place of accuracy as it provides a more detailed analysis of the data. The equation can be defined as
\begin{equation}
F\text{-}Measure = \frac{2TP}{2TP+FP+FN}.
\label{eq:F1}
\end{equation}

A downside to this measure is that it does not take into account \textit{TN}, a value that is required to create \textit{ROC} curves.
The \textit{MCC} uses all detection types to output a value between $-1$, which indicates total disagreement and $+1$, which indicates total agreement. A value of $0$ would be classed as a random prediction, and therefore both variables can be deemed independent. It can be provide a much more balanced evaluation of prediction than previous measurements, however it is not always possible to obtain all four detection types (i.e. \textit{TP}, \textit{FP}, \textit{FN}, \textit{TN}). The coefficient can be calculated by
\begin{equation}
\text{MCC} = \frac{ TP \times TN - FP \times FN } {\sqrt{ (TP + FP) ( TP + FN ) ( TN + FP ) ( TN + FN ) } }
\end{equation}

\subsection{Validation Techniques}
Two commonly used validation techniques in computer vision are \textit{n-fold} cross validation and leave-one-subject-out (LOSO). From our review, the evaluation system by different researchers reported in different validation techniques, where LOSO is more widely used. While some reported their results in both validation techniques \cite{liumain, davison2018samm}, and some only reported in LOSO \cite{kim2016micro, zheng2016relaxed, pfister2011differentiating}.
\section{Challenges}
Research on automated micro-expressions recognition using machine learning has witnessed good progress in recent years. A number of promising methods based on texture features, gradient features and optical flow features have been proposed. Many datasets was generated but lack of standardisation is indeed a great challenge. This section provides the challenges of the research in micro-expressions analysis into details.

\subsection{The effect of Spatial Temporal Settings in Data Collection}
Due to lack of communication between different research groups on experimental settings, the datasets are varied in resolution and frame rates. Some researchers \cite{li2013spontaneous, li2015reading} investigated on the effect of Temporal setting to micro-expression recognition. Using TIM \cite{pfister2011differentiating} to adjust the temporal settings is a well-known method in micro-expression analysis. However, there is a lack of thorough research in further investigating the implication of spatial-temporal changes for micro-expression recognition.

\begin{figure}
	\centering
	\includegraphics[scale=0.35]{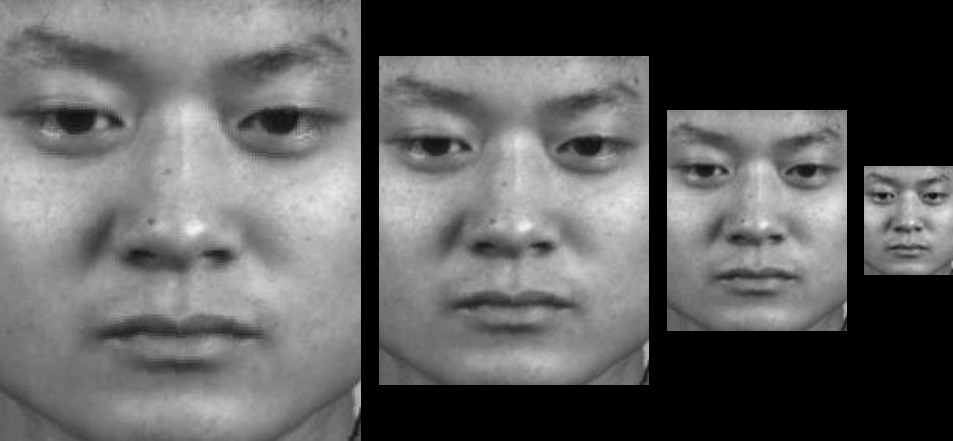}
	\caption{An example of different resolution by downscaling an image from CASME II dataset. From left to right: 100\% (Original resolution), 75\% of the original resolution, 50\% of the original resolution and 25\% of the original resolution.}
	\label{fig:downscale}
\end{figure}

\begin{figure}
	\centering
	\includegraphics[scale=0.4]{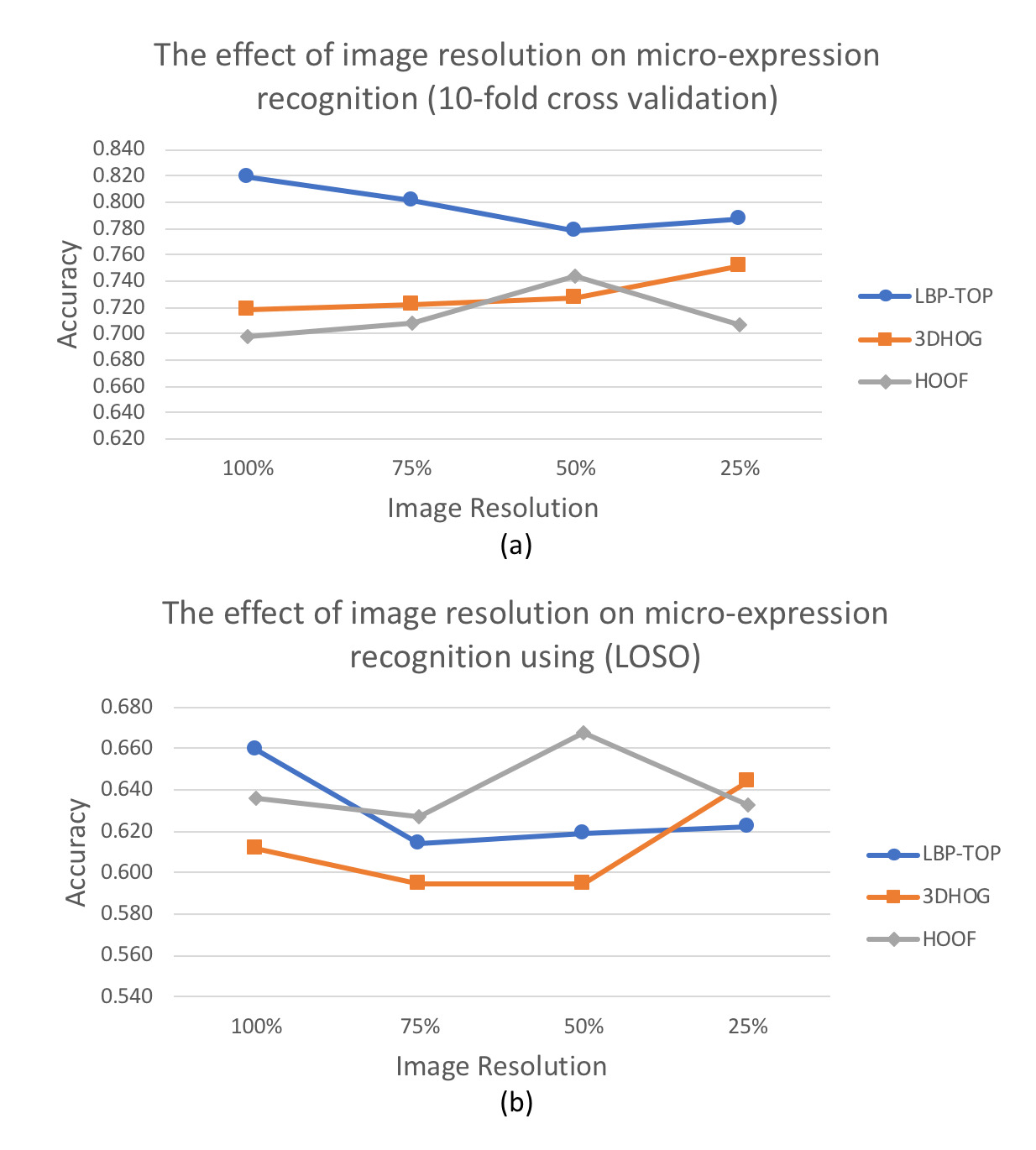}
	\caption{The effect of image resolution on micro-expression recognition using LBP-TOP, 3DHOG and HOOF on two different evaluation method: (a) 10-fold cross validation, and (b) LOSO.}
	\label{fig:resolution}
\end{figure}

We believe resolution plays an important role for features extraction. We downscale the CASME II dataset to four scales, 100\% (original resolution), 75\% of the original resolution, 50\% of the original resolution and 25\% of the original resolution, as depicted by Figure \ref{fig:downscale}. To address the research gap, we experiment this four resolutions with three feature types (LBP-TOP, 3DHOG and HOOF) with \textit{10-fold} cross validation and LOSO. To reduce the effect of learning algorithm, we used a standard SVM method as the classifier. Figure \ref{fig:resolution} compares the performance of the experiments.

From the observation, LBP-TOP performed better in high resolution images than 3DHOG and HOOF. It is noted that HOOF performed better when we downscale the resolution to 50\% and 3DHOG worked best at 25\%. These results showed LBP-TOP relied on spatial information (XY), but HOOF and 2DHOG are more dependent on temporal (XT and YT). The conventional methods are relies on feature descriptors and varies from one to another.

\subsection{Emotional Classes versus Objective Classes in Data Labelling}
A large focus on micro-expression research has been on the detection and recognition of emotion-based classed (i.e. discreet groups that micro-expression fit into during classification). Objective classes attempt to take away the potential bias of labelling difficult to distinguish micro-expressions into classes suited to a particular muscle movement pattern.

To date, SAMM~\cite{davison2018samm} is the only dataset that moves the focus from an emotional-based classification system, to an objective one, and is designed around analysing objective physical movement of muscles. Emotion classification requires the context of the situation for an interpreter to make
a meaningful interpretation. Most spontaneous micro-expression datasets have FACS ground truth labels
and estimated or predicted emotion. These have been annotated by an expert and self-reports written by
participants. In SAMM, Davison et al. \cite{davison2018samm} focused on objectiveness and did not report emotional classes in their dataset release. Due to this reason, it has not been widely experimented by other researchers. To address this issue, we introduced the emotional classes for SAMM in this paper.

SAMM has estimated emotional classes based on the AUs and the emotional stimuli presented to participants to allow for comparison with previous emotion class focused papers such as CASME II \cite{yan2014casme} and SMIC~\cite{li2013spontaneous}. The amount of clips in the SAMM dataset in each estimated emotion class can be seen in Table~\ref{tab:samm_emo_class}. Note that the categories are based around EMFACS labelling of reliable AUs to emotion~\cite{ekman1978facial}, so any that did not fit into these categories are placed in the \textquoteleft Other\textquoteright class.
\begin{table}
	\centering
	\caption{A breakdown of the number of clips categorised into estimated emotion classes for the SAMM dataset.}
	\label{tab:samm_emo_class}
	\begin{tabular}{|c|c|}
		\hline
		Estimated Emotion & Number of Clips \\ \hline
		Anger             & 57              \\ \hline
		Contempt          & 12              \\ \hline
		Disgust           & 9               \\ \hline
		Fear              & 8               \\ \hline
		Happiness         & 26              \\ \hline
		Sadness           & 6               \\ \hline
		Surprise          & 15              \\ \hline
		Other             & 26              \\ \hline
	\end{tabular}
\end{table}

To this end it can be argued that keeping classification to well-defined muscles (that cannot be changed or bias) is a more optimal solution to micro-expression recognition than discreet emotion classes. Further, Yan et al.~\cite{yan2014micro-expression} state that it’s inappropriate to categorise micro-expressions into emotion categories, and that using FACS AU research to inform the eventual emotional classification would be a more logical approach. In 2017, Davison et al. \cite{davison2017objective} proposed new objective classes based on FACS coding. They have coded the two state-of-the-art FACS-coded datasets into seven objective classes as illustrated in Table \ref{tab:AUCat}. This research could potentially be the new challenge for micro-expression research.

\begin{table}
	\centering
	\caption{Each class represents AUs that can be linked to emotion.}
	\label{tab:AUCat}
	\renewcommand{\arraystretch}{1.2}
	\begin{tabular}{|c|l|}
		\hline
		\multicolumn{1}{|l|}{Class} & Action Units \\ \hline
		I & AU6, AU12, AU6+AU12, AU6+AU7+AU12, AU7+AU12 \\ \hline
		II & \begin{tabular}[c]{@{}l@{}}AU1+AU2, AU5, AU25, AU1+AU2+AU25, AU25+AU26,\\ AU5+AU24\end{tabular} \\ \hline
		III & \begin{tabular}[c]{@{}l@{}}A23, AU4, AU4+AU7, AU4+AU5, AU4+AU5+AU7,\\ AU17+AU24, AU4+AU6+AU7, AU4+AU38\end{tabular} \\ \hline
		IV & \begin{tabular}[c]{@{}l@{}}AU10, AU9, AU4+AU9, AU4+AU40, AU4+AU5+AU40,\\ AU4+AU7+AU9, AU4 +AU9+AU17, AU4+AU7+AU10,\\ AU4+AU5+AU7+AU9, AU7+AU10\end{tabular} \\ \hline
		V & AU1, AU15, AU1+AU4, AU6+AU15, AU15+AU17 \\ \hline
		VI & AU1+AU2+AU4, AU20 \\ \hline
		VII & Others \\ \hline
	\end{tabular}
\end{table}

\subsection{Face Regions in Data Analysis}
\begin{figure}
	\centering
	\includegraphics[scale=0.5]{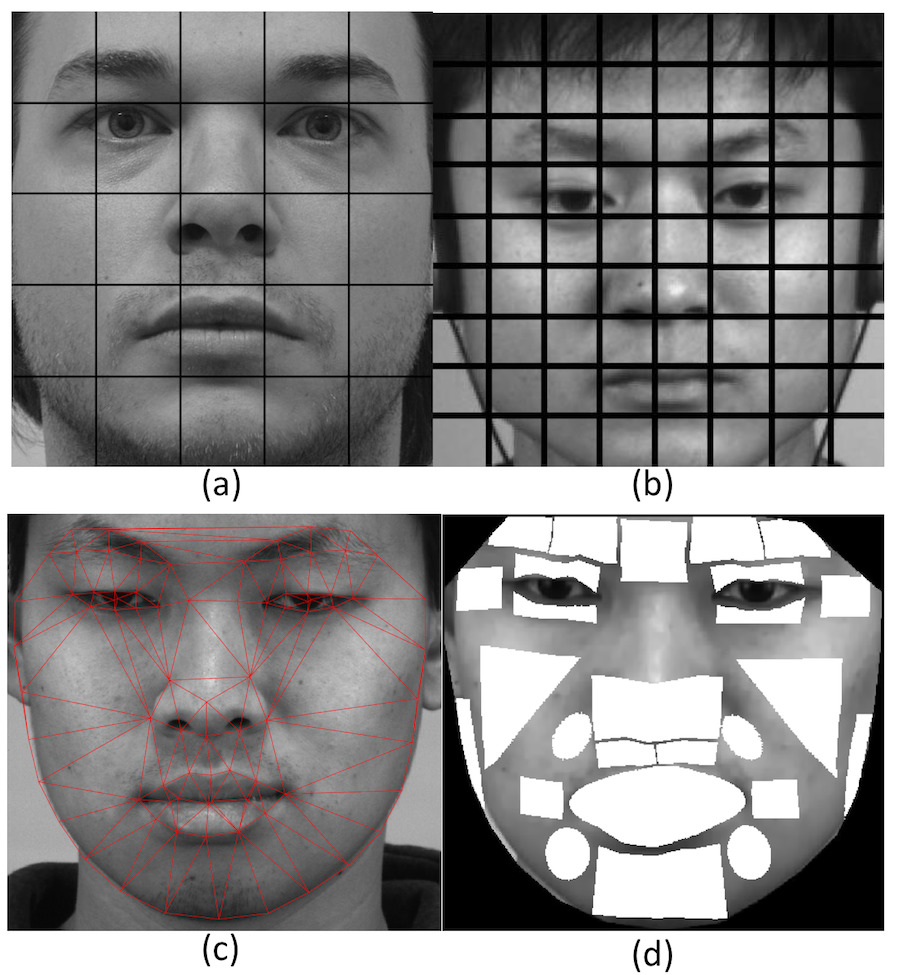}
	\caption{Illustration of face regions: (a) $5\times 5$ blocks, (b) $8\times 8$ blocks, (c) Delaunay triangulation, and (d) FACS-based regions.}
	\label{fig:face_region}
\end{figure}

Recent work on the micro-expressions recognition have provided promising results on successful detection techniques, however there is room for improvement. To begin detection, current approaches follow methods of extracting local feature information of the face by splitting the face into regions, as illustrated in Figure \ref{fig:face_region}. 

\begin{table*}[h]
	\centering
	\caption[Micro-movement method comparison.]{A Summary of the current micro-movement methods \cite{davison2016micro}. Each result metric changes depending on the method. * = true positives/recall, ** = area under curve.}
	\renewcommand{\arraystretch}{1.3}
	\begin{tabular}{llll} \hline
		Method & Feature & Dataset & Result \\ \hline
		Moilanen et al.~\cite{moilanen2014spotting} & LBP & CASME II/SMIC & 71\%* \\
		Shreve et al.~\cite{shreve2011macro} & Optical Strain & USF-HD & 74\%* \\
		Li et al.~\cite{li2015reading} & LBP & CASME II & 92.98\%** \\
		Xia et al.~\cite{xia2015spontaneous} & \begin{tabular}[c]{@{}l@{}}ASM Geometric\\ Deformation\end{tabular} & CASME/CASME & 92.08\%* \\
		Patel et al.~\cite{patel2015spatiotemporal} & Optical Flow & SMIC & 95\%** \\
		Davison et al. \cite{davison2015micro} & LBP/HOG & SAMM & 91.25\%* \\
		Davison et al. \cite{davison2017objective}  & 3D HOG & CASME II/SAMM & 68.04\%* \\ \hline
	\end{tabular}
	\label{tab:methodComp}
\end{table*}

The state of the art can be categorised into:
\begin{itemize}
	\item {\textit{Four quadrants}. Shreve et al.~\cite{shreve2014automatic} split the face into 4 quadrants and analyse each quarter as individual temporal sequences. The advantage of this method is that it is simple to analyse larger regions, however the information to retrieve from the areas are restricted to whether there was some form of movement in a more global area.}
	\item {\textit{$m \times n$ blocks}. Another method is to split the face into a specific number of blocks~\cite{yan2014casme, davison2014micro,davison2015micro}. The movement on the face is analysed locally, rather than a global representation of the whole face, and can focus on small changes in very specific temporal blocks. A disadvantage to this method is that it is computationally expensive to process the whole images as $m \times n$ blocks. It can also include features around the edge of the face, including hair, that do not relate to movement but could still effect the final feature vector. Figure \ref{fig:face_region}(a) and Figure \ref{fig:face_region}(b) illustrate the samples of block-based face regions.}
	\item {\textit{Delaunay triangulation}. Delaunay triangulation, as shown if Figure \ref{fig:face_region}(c), has also been used to form regions on just the face and can exclude hair and neck~\cite{lu2014delaunay}, however this approach can still extract areas of the face that would not be useful as a feature and adds further computational expense.}
	\item {\textit{FACS-based region}. A more recent and less researched approach is to use defined regions of interest (ROIs) to correspond with one or more FACS AUs~\cite{wang2014micro, wang2015micro}. These regions have more focus on local parts of the face that move due to muscle activation. Some examples of ROI selection for micro-expression recognition and detection include discriminative response map fitting~\cite{liumain}, Delaunay triangulation~\cite{lu2014delaunay} and facial landmark based region selection~\cite{patel2015spatiotemporal}. Unfortunately, currently defined regions do not cover all AUs and miss some potentially important movements such as AU5 (Upper Lid Raiser), AU23 (Lip Tightener) and AU31 (Jaw Clencher). To overcome the problem, Davison et al. \cite{davison2017objective} proposed FACS-based regions to improve local feature representation by disregarding face region that do not contribute to facial muscle movements. The defined region is presented in Figure \ref{fig:face_region}(b).}
\end{itemize}

Figure \ref{fig:face_region} compares different face region splitting methods. Due to FACS-based region is more relevant to facial muscle movements and suitable for AUs detection, more research should be focusing on FACS-based region than split the face into $m\times n$ blocks.

\subsection{Deep Learning versus Conventional Approaches}

The pipeline of conventional micro-expression recognition approach is very similar to macro-expressions in terms of preprocessing techniques, hand-crafted features and, if applicable, machine learning classification. However, geometric feature-based methods are rarely used as tracking feature points on a face that barely moves will not produce good results. Instead, appearance-based features are primarily used to attempt to describe the micro-movement or train machine learning to classify micro-expressions into classes. 

Spatial temporal settings during data collection, preprocessing stage of dataset including face alignment and face regions split, feature extraction methods and the type of classifiers are the main factors for conventional approaches. Moving forward, end-to-end solution that is capable of handling these issues is required. Deep learning approaches have yet to have much impact on micro-expression analysis, however to ensure a rounded review of current techniques we shall provide a preliminary study on deep learning and its applications to micro-expression.

As the temporal nature of micro-expressions are a key feature to understand, modern video-analysis technique, namely 3D convolutional neural networks (3D ConvNets)~\cite{tran2015learning}, may be used to exploit the temporal dimension. This network expands on the typical 2D convolutional neural network (CNN) by using 3$\times$3$\times$3 convolutional kernels where the third dimension is in the temporal domain (frames in a video). It was originally used for analysis for action recognition, however it can be expanded for any other video-analysis task easily. Using the deconvolution method described by Zeiler and Fergus~\cite{zeiler2014visualizing}, Tran et al.~\cite{tran2015learning} was able to show that the features extracted from the 3D ConvNet focuses on the appearance of the first few frames and then tracks salient motion over the next frames. The key difference in using 2D ConvNets is the ability to extract and learn from features from both motion and appearance.

With minimal data available to train from, deep learning methods have a much more difficult time in learning meaningful patterns~\cite{deng2014deep}. When independent test samples were used for validation, the model showed that further investigation is required for deep learning with micro-expression to be effective, including the use of more data. The biggest disadvantage to using video-data is not being able to load such large amounts of data into memory, even on GPUs that have 12GB of on-board memory. This leads to the minimisation of the batch size and reduction of resolution to allow for training to proceed. Further ways of being able to handle micro-expression data without having to reduce the amount of data available would be vital to retaining the discriminative information required for micro-expression analysis. Further, the time required to train the model shows the challenge of the ability to train long video-based deep learning methods.

\subsection{Spotting Micro-movement on the Face}
Micro-expressions analysis tends to focus on the emotion recognition, meaning assumptions are commonly made. Focusing on micro-facial movements, which describe the facial muscle activations, removes these assumptions \cite{davison2018samm}. Table \ref{tab:methodComp} summarised the benchmark publications in micro-movement \cite{davison2016micro}, it is notably less publication when compared to micro-expressions recognition. However, this is equally important as not all the AUs are linked with emotional context. Future challenge will be focusing on spotting micro-movements on long videos. There are limited datasets provide long video clips. One of the dataset provides is SAMM, where the researchers can received it by posting a physical hard disk to obtain the full dataset (700GB).

\subsection{Standardisation of Metrics}
We recommend the researchers to standardised the performance metrics that they used in evaluation. As the majority of datasets are inbalanced \cite{le2014spontaneous}, reporting the result in \textit{F-Measure} (or \textit{F1-Score}) seems to be the best option. Using the conventional \textit{Accuracy} measure may result in a bias towards classes with large number of samples, hence overestimating the capability of the evaluated method. F-Measure micro-average across the whole dataset and is computed based on the total true positives, false negatives and false positives, across 10-fold cross validation and Leave-one-subject-out (LOSO). 

Due to each dataset with small micro-expression samples, the researchers are encourage to use more datasets for their experiment. For cross datasets evaluation, unweighted average recall (UAR) and weighted average recall (WAR) are recommended as these measurements were shown promising in speech emotion recognition \cite{schuller2010cross}. WAR is defined as number of correctly classified samples divided by the total number of samples, while UAR is defined as sum of accuracy of each class divided by the number of classes without considerations of samples per class. To obtain the overall scores, the results from all the folds are averaged. These metrics had been recommended in the First Micro-expressions Grand Challenge Workshop in conjunction with Face and Gesture 2018 Conference.

\subsection{Real-world Implementation}
 For implementation of the micro-expressions recognition in real-world, the challenges to be addressed include:
\begin{itemize}
	\item \textbf{Cross-Cultural Analysis} Micro-facial expressions occur when people attempt to hide their true emotion, and so the possibility of how well some cultures manage this suppression would be interesting to learn. By using software to detect micro-expressions across cultures, the results of different suppression of emotion can be studied. Therefore people in East Asian cultures could be different from Western cultures, which can be analysed to find any correlation between the psychological studies and automated micro-expressions recognition. Something to note in this type of investigation would be to ensure the different participants originate and live in their respective countries, as people living with different cultures for a long time may not exhibit the same behaviour.
\end{itemize}
\begin{itemize}
	\item \textbf{Dataset Improvements}. Further work can be done to improve micro-movement datasets. Firstly, more datasets or expanding previous sets would be a simple improvement that can help move the research forward faster. Secondly, a standard procedure on how to maximise the amount of micro-movements induced spontaneously in laboratory controlled experiments would be beneficial. If collaboration between established datasets and researchers from psychology occurred, dataset creation would be more consistent. As using human participants is required, and emotions are induced, ethical concerns are always going to play a part in future studies of this kind. Any work moving forward must take into account these concerns and draw from previous experiments to ensure no harm will come to the psychological welfare of participants.
\end{itemize}
\begin{itemize}
	\item \textbf{Real-Time Micro-Facial Expressions Recognition}. To be able to implement any form of micro-movement detection system into a real-world scenario, it must perform the processes required in real-time (or near to real-time). As the accuracy of facial expression analysis is already quite high, transitioning to real-time has already produced decent results. However there is currently no known systems that is able to detect micro-expressions.
\end{itemize}

The accuracy of many state-of-the-art methods is still too low to be deployed effectively in a real-world environment. The progress in research of micro-expressions recognition can aid in the paradigm shift in affect computing for real-world applications in psychology, health study and security control.

\section{Conclusion}
We have presented a comprehensive review on datasets, features and metrics for micro-expressions analysis. The ultimate goal of this paper is to provide new insights and recommendations to advancing the micro-expression analysis research. We have provided a good guidelines for beginners and a detailed challenges and recommendations for those who are already working in this area. In addition, we contribute to the research by addressing the effect of resolutions on different feature types and introducing the new emotional classes for SAMM.

To summarise, the future direction to advance automated micro-expression recognition should take into consideration on how the dataset is capture (spatial temporal settings), labeling of the dataset based on Action Unit based objective classes, FACS-based face regions for better localisation, end-to-end solution using deep learning, fair evaluation using standardised metrics (ideally F1-Score and MCC) and LOSO as the validation technique. More importantly, the openness and better communication within the research communities are crucial to crowd-source the data labelling and using the standard evaluation system.

As micro-expression recognition is still in its infancy when compared to the macro-expression, it requires combined efforts from multidisciplinary (including psychology, computer science, physiology, engineer and policy maker) to achieve reliable results for practical real-world application. A controversial point is whether or not it should be allowed to detect these micro-expressions, as the theory behind it states that the person attempting to conceal their emotion experience these movements involuntarily and likely unknowingly. If we are able to detect them with high accuracy, then we are effectively robbing a person of being able to hide something that is private to them. From an ethical point of view, knowing when someone is being deceptive would be advantageous but takes away the freedom you had in your emotions.

%



\ifCLASSOPTIONcompsoc
  \section*{Acknowledgments}
\else
  \section*{Acknowledgment}
\fi

The authors would like to thank Prof. Xiaolan Fu of The Institute of Psychology, Chinese
Academy of Sciences for offering CASME II micro-expression database for this research. The authors would like to thank The Royal Society Industry Fellowship (Grant number: IF160006, awarded to Dr. Moi Hoon Yap).

\ifCLASSOPTIONcaptionsoff
  \newpage
\fi



\bibliographystyle{IEEEtran}
\bibliography{IEEEabrv}
\end{document}